\documentclass{proc}
\usepackage[dvips]{graphicx}
\usepackage{amssymb,amsmath}
\usepackage{stmaryrd}               
\usepackage{listings}     

\makeatletter
\def\@maketitle{%
  \vbox to 2.3in{%
    \hsize\textwidth
    \linewidth\hsize
    \vspace*{1.5cm}
    \centering
    {\bfseries\LARGE \@title \par}
    \vskip 2em
    {\large \begin{tabular}[t]{c}\@author \end{tabular}\par}
    \vfill}    \vspace*{1.0cm}
}
\renewcommand\section{\@startsection {section}{1}{\z@}%
     {.7\baselineskip plus\baselineskip}{.5\baselineskip}
                                   {\normalfont\Large\bfseries}}
\renewcommand\section{\@startsection {section}{1}{\z@}%
      {.5\baselineskip\@plus.7\baselineskip}{.3\baselineskip}%
                                   {\normalfont\Large\bfseries}}
\renewcommand\subsection{\@startsection{subsection}{2}{\z@}%
       {.5\baselineskip\@plus.7\baselineskip}{.3\baselineskip}%
                                   {\normalfont\large\bfseries}}
\renewcommand\subsubsection{\@startsection{subsubsection}{3}{\z@}%
      {.5\baselineskip\@plus.7\baselineskip}{.3\baselineskip}%
                                     {\normalfont\normalsize\bfseries}}
\makeatother

\renewenvironment{abstract}%
  {\normalfont
    \list{}{\labelwidth0pt
      \leftmargin0pt \rightmargin\leftmargin
      \listparindent\parindent \itemindent0pt
      \parsep0pt
      
    }%
    \item[\hskip\labelsep\bfseries\abstractname\enspace --] \itshape%
}{%
  \endlist}

\newcommand{\keywordsname}{Keywords}
\newenvironment{keywords}%
  {\normalfont
    \list{}{\labelwidth0pt
      \leftmargin0pt \rightmargin\leftmargin
      \listparindent\parindent \itemindent0pt
      \parsep0pt
      }%
    \item[\hskip\labelsep\bfseries\keywordsname:]}{\endlist}

\begin{document}

\title{Threat assessment of a possible Vehicle-Born \\ Improvised Explosive Device
using DSmT}

\author{
{\bf Jean~Dezert}\\
French Aerospace Lab.\\
ONERA/DTIM/SIF\\
29 Av. de la Div. Leclerc\\
92320 Ch\^{a}tillon, France.\\
\underbar{jean.dezert@onera.fr}\\
\and
{\bf Florentin Smarandache}\\
ENSIETA\\
$\text{E}^3\text{I}^2\text{-EA3876}$ Laboratory\\
2 rue Fran\c{c}ois Verny\\
29806 Brest Cedex 9, France.\\
\underbar{Florentin.Smarandache@ensieta.fr}
}

\date{}

\maketitle

\begin{abstract}
This paper presents the solution about the threat of a VBIED (Vehicle-Born Improvised Explosive Device) obtained with the DSmT (Dezert-Smarandache Theory). This problem has been proposed recently to the authors by Simon Maskell and John Lavery as a typical illustrative example to try to compare the different approaches for dealing with uncertainty for decision-making support. The purpose of this paper is to show in details how a solid justified solution can be obtained from DSmT approach and its fusion rules thanks to a proper modeling of the belief functions involved in this problem. 
\end{abstract}


\section{The VBIED problem}
\noindent

\begin{itemize}
\item {\bf{Concern}}: VBIED (Vehicle-Born Improvised Explosive Device) attack on an administrative building $B$
\item {\bf{Prior information}}: We consider an Individual $A$ under surveillance due to previous unstable behavior who drives customized white Toyota (WT) vehicle.
 \item {\bf{Observation done at time}} $t-10\ \text{min}$: From a video sensor on road that leads to building $B$ 10 min ago, one has observed a White Toyota 200m from the building $B$ traveling in normal traffic flow toward building $B$. We consider the following two sources of information based on this video observation available at time $t-10\ \text{min}$:
 
 \begin{itemize}
 \item {\bf{Source 1}}: An Analyst 1 with 10 years experience analyses the video and concludes that individual $A$ is now probably near building $B$.
 \item {\bf{Source 2}}: An Automatic Number Plate Recognition (ANPR) system analyzing same video outputs 30\% probability that the vehicle is individual $A$'s white Toyota.
\end{itemize}

 \item {\bf{Observation done at time}} $t-5\ \text{min}$: From a video sensor on road 15km from building $B$ 5 min ago one gets a video that indicates a white Toyota with some resemblance to individual $A$'s white Toyota. We consider the following thrid source of information based on this video observation available at time $t-5\ \text{min}$:
 \begin{itemize}
 \item {\bf{Source 3}}: An Analyst 2 (new in post) analyses this video and concludes that it is improbable that individual $A$ is near building $B$.
 \end{itemize}

 \item {\bf{Question 1}}: ÊShould building B be evacuated?

 \item {\bf{Question 2}}: ÊIs experience (Analyst 1) more valuable than physics (the
ANPR system) combined with inexperience (Analyst 2)? How do we model that?
 
\end{itemize}

\noindent
NOTE: Deception (e.g., individual $A$ using different car, false number plates, etc.) and biasing (on the part of the analysts) are often a part of reality, but they are not part of this example.
 
\section{Modeling the VBIED problem}
 
Before applying DSmT fusion techniques to solve this VBIED problem it is important to model the problem in the framework of belief functions.\medskip

\subsection{Marginal frames with their models}

The marginal frames involved in this problem are:

\begin{itemize}
\item Frame related with individuals: 
\begin{multline*}
\Theta_1=\{A=\text{Suspicious person},\bar{A}=\text{not $A$}\}
\end{multline*}
\item Frame related with the vehicle: 
$$\Theta_2=\{V=\text{White Toyota Vehicle},\bar{V}=\text{not $V$}\}$$
\item Frame related with the position of a driver of a car w.r.t the given building $B$: 
$$\Theta_3=\{B=\text{near building},\bar{B}=\text{not $B$}\}$$
\end{itemize}

\noindent
The underlying models of marginal frames are based on the following very reasonable assumptions:
\begin{itemize}
\item {\bf{Assumption 1}}: We assume naturally $A\cap \bar{A}=\emptyset$ (avoiding Shršdinger's cat paradox). If working only with the frame of people $\Theta_{3}$, the marginal bba's must be defined on the power-set
 $$2^{\Theta_1} = \{\emptyset_1,A,\bar{A},A\cup\bar{A}\}$$
\end{itemize}

\begin{itemize}
\item  {\bf{Assumption 2}}: We assume also that $V\cap \bar{V}=\emptyset$ so that the marginal bba (if needed) must be defined on the power-set
$$2^{\Theta_2} = \{\emptyset_2,V,\bar{V},V\cup\bar{V}\}$$
\end{itemize}

\begin{itemize}
\item  {\bf{Assumption 3}}: We assume also that $B\cap \bar{B}=\emptyset$ so that the marginal bba (if needed) must be defined on the power-set
$$2^{\Theta_3} = \{\emptyset_3,B,\bar{B},B\cup\bar{B}\}$$
This modeling is disputable since the notion of  closeness/''near''  is not clearly defined and we could prefer to work on
$$D^{\Theta_3} = \{\emptyset_3,B\cap\bar{B},B,\bar{B},B\cup\bar{B}\}$$
\end{itemize}

The emptyset elements have been indexed by the index of the frame they are referring to for notation convenience and avoiding confusion.\\

\subsection{Joint frame and its model}

Since we need to work with all aspects of available information, we need to define a common joint frame to express all what we have from different sources of information.
The easiest way for defining the joint frame, denoted $\Theta$, is to consider the classical Cartesian (cross) product space and to work with propositions (a Lindenbaum-Tarski algebra of propositions) since one has a correspondence between sets and propositions \cite{Shafer_1976,DSmTBook1-3}, i.e.
$$\Theta= \Theta_{1}\times \Theta_{2} \times \Theta_{3}$$
\noindent
which consists of the following 8 triplets elements
\begin{align*}
\Theta=\{&\theta_1=(\bar{A},\bar{V},\bar{B}), \theta_2=(A,\bar{V},\bar{B}),\\
                &\theta_3=(\bar{A},V,\bar{B}), \theta_4=(A,V,\bar{B}),\\
                &\theta_5=(\bar{A},\bar{V},B), \theta_6=(A,\bar{V},B),\\
                &\theta_7=(\bar{A},V,B), \theta_8=(A,V,B) \}
\end{align*}

We define the union $\cup$ , intersection $\cap$ as componentwise operators in the following way:
$$(x_{1},x_{2},x_{3})\cup(y_{1},y_{2},y_{3}) \triangleq (x_{1}\cup y_{1},x_{2}\cup y_{2},x_{3}\cup y_{3})$$
$$(x_{1},x_{2},x_{3})\cap(y_{1},y_{2},y_{3}) \triangleq (x_{1}\cap y_{1},x_{2}\cap y_{2},x_{3}\cap y_{3})$$
\noindent
The complement $\bar{X}$ of $X$ is defined in the usual way by
$$\bar{X}=\overline{(x_{1},x_{2},x_{3})} \triangleq I_t \setminus \{X\}$$
\noindent
where $I_t$ is the total ignorance (i.e. the whole space of solutions) which corresponds to the
maximal element defined by $I_{t} = (I_{t1}, I_{t2}, I_{t3})$, where $I_{ti}$ is the maximal (ignorance) of $\Theta_i$, $i=1,2,3$. The minimum element (absolute empty proposition) is 
$\emptyset = (\emptyset_{1}, \emptyset_{2}, \emptyset_{3})$, where $\emptyset_{i}$ is the minimum element (empty proposition) of $\Theta_i$. We also define a relative minimum element in $S^{\Theta_{1}\times \Theta_{2} \times \Theta_{3}}$ as follows: $\emptyset_r = (x,y,z)$, where at least one of the components $x$, $y$, or $z$ is a minimal element in its respective frame $\Theta_i$. A general relative minimum element $\emptyset_{gr}$ is defined as the union/join of all relative minima (including the absolute minimum element). Similarly to the relative and general relative minimum we can define a relative maximum and a general relative maximum, where the empty set in the above definitions is replaced by the total ignorance. Whence the super-power set $(S^{\Theta}, \cap,\cup,\bar{\quad},\emptyset , I_t)$ is equivalent to Lindenbaum-Tarski algebra of propositions. \\

\noindent
For example, if we consider $\Theta_{1}=\{x_{1},x_{2}\}$ and $\Theta_{2}=\{y_{1},y_{2}\}$ satisfying both Shafer's model, then $\Theta=\Theta_{1}\times \Theta_{2}=\{(x_{1},y_{1}),(x_{1},y_{2}),(x_{2},y_{1}),(x_{2},y_{2})\}$, and one has:
\begin{align*}
\emptyset & =(\emptyset_{1},\emptyset_{2})\\
\emptyset_{r1}&=(\emptyset_{1},y_{1})\\
\emptyset_{r2}&=(\emptyset_{1},y_{2})\\
\emptyset_{r3}&=(\emptyset_{1},y_{1}\cup y_{2})\\
\emptyset_{r4}&=(x_{1},\emptyset_{2})\\
\emptyset_{r5}&=(x_{2},\emptyset_{2})\\
\emptyset_{r6}&=(x_{1}\cup x_{2},\emptyset_{2})
\end{align*}
\noindent
and thus
$$\emptyset_{gr} = \emptyset\cup \emptyset_{r1}\cup \emptyset_{r2}\cup \ldots \cup \emptyset_{r6}$$

Based on definition of joint frame $\Theta$ with operations on its elements, we need to choose its underlying model (Shafer's, free or hybrid model) to define its fusion space where the bba's will be defined on. According to the definition of absolute and relative minimal elements, we then assume for the given VBIED problem  that $\Theta$ satisfies Shafer's model, i.e. all (triplets) elements $\theta_{i} \in \Theta$ are exclusive, so that the bba's of sources will be defined on the classical power-set $2^\Theta$.

\subsection{Supporting hypotheses for decision}

In the VBIED problem the main question (Q1) is related with the security of people in the building $B$. The potential danger related with this building is of course $\theta_8=(A,V,B)$ i.e. the presence of $A$ in his/her car $V$ near the building $B$. This is however and unfortunately not the only origin of the danger since the threat can also come from the possible presence of $V$ (possible A's improvised explosive vehicle) parked near the building $B$ even if $A$ has left his/her car and is not himself/herself near the building. This second origin of danger is represented by $\theta_7=(\bar{A},V,B)$. There exists also a third origin of the danger represented by $\theta_6=(A,\bar{V},B)$ which reflects the possibility to have $A$ near the building without $V$ car. $\theta_6$ is also dangerous for the building $B$ since $A$ can try to commit a suicidal terrorism attack as human bomb against the building. Therefore based on these three sources of potential danger, the most reasonable/prudent supporting hypothesis for decision-making is consider
$$\theta_6\cup \theta_7 \cup \theta_8 = (A,\bar{V},B)\cup (\bar{A},V,B)\cup (A,V,B)$$

If we assume that the danger is mostly due to presence of $A$'s vehicle containing possibly a high charge of explosive near the building $B$ rather than the human bomb attack, then one can prefer to consider only  the following hypothesis for decision-making support evaluation
$$\theta_7 \cup \theta_8 = (\bar{A},V,B)\cup (A,V,B)$$

Finally if we are more {\it{optimistic}}, we can consider that the real danger occurs if and only if $A$ drives $V$ near the building $B$ and therefore one could consider only the supporting hypothesis $\theta_8 = (A,V,B)$ for the danger in the decision-making support evaluation.\\

In the sequel, we adopt the worst scenario (we take the most prudent choice) and we consider all three origins of potential danger. Thus we will take $\theta_6\cup \theta_7 \cup \theta_8$ as cautious/prudent supporting hypothesis for decision-making.\\

Thepropositions $\theta_6 \cup \theta_8 = (A,\bar{V},B)\cup (A,V,B)$ and $\theta_6 \cup \theta_7 =  (A,\bar{V},B)\cup (\bar{A},V,B)$ represent also a potential danger and could serve as decision-support hypotheses also, and their imprecise probabilities can be evaluate easily following analysis presented in the sequel. They have not been reported in this paper to keep it at a reasonable size.

\subsection{Choice of bba's of sources}

Let's define first the bba of each source without regard to what could be their reliability and importance in the fusion process. Reliability and importance will be examined in details in next section.

\begin{itemize}
\item  {\bf{Bba related with source 0}} (prior information): The prior information states that the suspect $A$ drives a white Toyota, and nothing is state about the prior information with respect to his location, so that we must consider the bba's representing the prior information as

\begin{align*}
m_{0}(\theta_{4}\cup \theta_{8}) & = m_{0}( (A,V,\bar{B}) \cup (A,V,B)) \\
& =  m_{0}((A,V,B\cup\bar{B}))\\
& =1
\end{align*}
\end{itemize}

\begin{itemize}
\item  {\bf{Bba related with source 1}} (Analyst 1 with 10 years experience): The source 1 reports that the suspect $A$ is probably now near the building $B$. This source however doesn't report explicitly that the suspect $A$ is still with its white Toyota car or not. So the fair way to model this report when working on $\Theta$ is to commit a high mass of belief to the element $\theta_{6}\cup \theta_{8}$, that is
\begin{align*}
m_{1}(\theta_{6}\cup \theta_{8}) & = m_{1}( (A,\bar{V},B) \cup (A,V,B)) \\
& =  m_{1}((A,V\cup\bar{V},B))\\
& =0.75
\end{align*}
\noindent
and to commit the uncommitted mass to $I_{t}$ based on the principle of minimum of specificity, so that
$$m_{1}(\theta_{6}\cup \theta_{8})=0.75 \quad \text{and} \quad m_{1}(I_{t})=0.25$$
\end{itemize}

\begin{itemize}
\item  {\bf{Bba related with source 2}} (ANPR system): The source 3 reports 30\% probability that the vehicle is individual $A$'s wite Toyota. Nothing is reported on the position information.  The information provided by this source corresponds actually to incomplete probabilistic information. Indeed, when working on $\Theta_1\times \Theta_2$, what we only know is that $P\{(A,V)\}=0.3$ and $P\{(\bar{A},V)\cup (A,\bar{V})\cup (\bar{A},\bar{V})\}=0.7$ (from additivity axiom of probability theory) and thus the bba $m_2(.)$ we must choose on $\Theta_1\times \Theta_2\times \Theta_3$ has to be compatible with this incomplete probabilistic information, i.e. the projection $m_2'(.)\triangleq m_2^{\downarrow\Theta_1\times \Theta_2}(.)$ of $m_2(.)$ on $\Theta_1\times \Theta_2$ must satisfy the following constraints on belief and plausibility functions
$$Bel'((A,V))=0.3$$
$$Bel'((\bar{A},V)\cup(A,\bar{V})\cup (\bar{A},\bar{V}))=0.7$$
\noindent
and also
$$Pl'((A,V))=0.3$$
$$Pl'((\bar{A},V)\cup (A,\bar{V})\cup (\bar{A},\bar{V}))=0.7$$

\noindent
because belief and plausibility correspond to lower and upper bounds of probability measure \cite{Shafer_1976}.
So it is easy to verify that the following bba $m_2'(.)$ satisfy these constraints because the elements of the frame $\Theta_1\times \Theta_2$ are exclusive:
$$m_2'((A,V))=0.3$$
$$m_2'((\bar{A},V)\cup (A,\bar{V})\cup (\bar{A},\bar{V}))=0.7$$

\noindent
We can then extend $m_2'(.)$ into $\Theta_1\times \Theta_2\times \Theta_3$ using the minimum specificity principle (i.e. take the vacuous extension of $m_2'(.)$) to get the bba $m_2(.)$ that we need to solve the VBIED problem. That is $m_2(.)=m_2'^{\uparrow \Theta_1\times \Theta_2\times \Theta_3}(.)$ with

$$m_2((A,V,B\cup\bar{B}))=0.3$$
$$m_2((\bar{A},V,B\cup\bar{B})\cup (A,\bar{V},B\cup\bar{B})\cup (\bar{A},\bar{V},B\cup\bar{B}))=0.7$$

\noindent
or equivalently
$$m_2(\theta_4\cup\theta_8)=0.3$$
$$m_2(\overline{\theta_4\cup\theta_8})=m_2(\theta_1\cup\theta_2\cup\theta_3\cup\theta_5\cup\theta_6\cup\theta_7)=0.7$$

\end{itemize}

\begin{itemize}
\item  {\bf{Bba related with source 3}} (Analyst 3 with no experience): The source 3 reports that it is improbable that the suspect $A$ is near the building $B$. This source however doesn't report explicitly that the suspect $A$ is still with its white Toyota car or not. So the fair way to model this report when working on $\Theta$ is to commit a low mass of belief to the element $\theta_{6}\cup \theta_{8}$, that is
\begin{align*}
m_{3}(\theta_{6}\cup \theta_{8}) & = m_{3}( (A,\bar{V},B) \cup (A,V,B)) \\
& =  m_{3}((A,V\cup\bar{V},B))\\
& =0.25
\end{align*}
\noindent
and to commit the uncommitted mass to $I_{t}$ based on the principle of minimum of specificity, so that
$$m_{3}(\theta_{6}\cup \theta_{8})=0.25 \quad \text{and} \quad m_{3}(I_{t})=0.75$$
\end{itemize}

\subsection{Reliability of sources}

Let's identify what is known about the reliability of sources and information:
\begin{itemize}
\item {\bf{Reliability of prior information}}: it is (implicitly) supposed that the prior information is 100\% reliable that is ''Suspect $A$ drives a white Toyota'' which corresponds to the element $(A,V,B)\cup (A,V,\bar{B})$. So we can take the reliability factor of prior information as $\alpha_{0}=1$. If one considers the priori information highly reliable (but not totally reliable) then one could take $\alpha_{0}=0.9$ so that $m_{0}(.)$ would be
$$m_{0}(\theta_{4}\cup \theta_{8})=0.9 \quad \text{and} \quad m_{0}(I_{t})=0.1$$
\item {\bf{Reliability of source 1}}:  One knows that Analyst \# 1 has 10 years experience, so we must consider him/her having a good reliability (say greater than 75\%) or to be less precise we can just assign to him a qualitative reliability factor with minimal number of labels in $\{L_{1}=\text{not good},L_{2}=\text{good}\}$. Here we should choose $\alpha_{1}=L_{2}$. As first approximation, we can consider $\alpha_{1}=1$.
 \item {\bf{Reliability of source 2}}: No information about the reliability of ANPR system is explicitly given. We may consider that if such device is used it is because it is also considered as a valuable tool and thus we assume it has a good reliability too, that is $\alpha_{2}=1$. If we want to be more prudent we should consider the reliability factor of this source as totally unknown and thus we should take it as very imprecise with $\alpha_{2}=[0,1]$ (or qualitatively as $\alpha_{2}=[L_{0},L_{3}]$). If we are more optimistic and consider ANPR system as reliable enough, we could take $\alpha_{2}$ a bit more precise with $\alpha_{2}=[0.75,1]$ (i.e. $\alpha_{2}\geq 0.75$) or just qualitatively as
 $\alpha_{2}=L_{2}$.
  \item {\bf{Reliability of source 3}}: It is said explicitly that Analyst 2 is new in post, which means that Analyst 2 has no great experience and it can be inferred logically that it is less reliable than Analyst 1 so that we must choose $\alpha_{3} < \alpha_{1}$. But we can also have a very young brillant analyst who perform very well too with respect to the older Analyst 1. So to be more cautious/prudent, we should also consider the case of unknown reliability factor $\alpha_{3}$ by taking qualitatively $\alpha_{3}=[L_{0},L_{3}]$ or quantitatively by taking $\alpha_{3}$ as a very imprecise value that is $\alpha_{3}=[0,1]$.
\end{itemize}

\subsection{Importance of sources}

\noindent
Not that much is explicitly said about the importance of the sources of information in the VBIED problem statement, but the fact that Analyst 1 has ten years experience and Analyst 2 is new in post, so that it seems logical to choose as importance factor $\beta_{1} > \beta_{3}$. The importances discounting factors have been introduced and presented by the authors in \cite{Dezert_2010b,FSJDJMT2010}. As a prudent attitude we could choose also $\beta_{0}=[0,1]=[L_{0},L_{3}]$ and $\beta_{2}=[0,1]=[L_{0},L_{3}]$ (vey imprecise values). If we consider that the prior information and the source 2 (ANPR) have the same importance, we could just take $\beta_{0}=\beta_{1}=1$ to make derivations easier and adopt a more optimistic point of view\footnote{Of course the importance discounting factors can also be chosen approximatively from exogenous information upon the desiderata of the fusion system designer. This question is out of the scope of this paper.}.

\section{Solution of VBIED problem}

We apply PCR5 and PCR6 fusion rules developed originally in the DSmT framework to get the solution of the VBIED problem. PCR5 has been developed by the authors in \cite{DSmTBook1-3}, Vol.2, and PCR6 is a variant of PCR5 proposed by Arnaud Martin and Christophe Osswald in \cite{MartinDSmTBook2}. Several codes for using PCR5 and PCR6 have been proposed in the literature for example in \cite{MartinDSmTBook2,Dambreville_2010,FSJDJMT2010} and are available to the authors upon request.\\

Two cases are explored depending on the taking into account or not of the reliability and the importance of sources in the fusion process.
To simplify the presentation of the results we denote the focal elements involved in this VBIED problem as:
\begin{align*}
f_1& \triangleq \theta_{4}\cup \theta_{8}\\
f_2& \triangleq \theta_{6}\cup \theta_{8}\\
f_3& \triangleq \theta_1\cup\theta_2\cup\theta_3\cup\theta_5\cup\theta_6\cup\theta_7=\overline{\theta_{4}\cup \theta_{8}}\\
f_4& \triangleq I_t =\theta_1\cup\theta_2\cup\theta_3\cup\theta_4\cup\theta_5\cup\theta_6\cup\theta_7\cup\theta_8
\end{align*}

Only these focal elements are involved in inputs of the problem and we recall the two questions that we must answer:\medskip

 \noindent
 {\bf{Question 1 (Q1)}}: ÊShould building B be evacuated?
 
 The question 1 must be answered by analyzing the level of belief and plausibility committed in the propositions supporting $B$ through the fusion process.\\

 \noindent
 {\bf{Question 2 (Q2)}}: ÊIs experience (Analyst 1) more valuable than physics (the
ANPR system) combined with inexperience (Analyst 2)? How do we model that?

The question 2 must be answered by analyzing and comparing the results of the fusion $m_1\oplus m_3$ (or eventually $m_0\oplus m_1\oplus m_3$) with respect to $m_2$ only (resp. $m_0\oplus m_2$).

\subsection{Without reliability and importance}

We provide here the solutions of the VBIED problem with direct PCR5 and PCR6 fusion of the sources for different qualitative inputs summarized in the tables below. We also present the result of DSmP probabilistic transformation \cite{DSmTBook1-3} (Vol.3, Chap. 3) of resulting bba's to get and approximate probability measure of elements of $\Theta$. No importance and reliability discounting has been applied since in this section, we consider that all sources have same importances and same reliabilities.\\


\noindent
{\bf{Example 1}}: We take the bba's described in section 2.3, that is

 \begin{table}[!h]
 \small
\centering
 \begin{tabular}{|l|c|c|c|c|}
    \hline
    focal element     & $m_{0}(.)$  & $m_{1}(.)$ & $m_{2}(.)$ & $m_{3}(.)$ \\
    \hline  
    $\theta_4\cup\theta_8$  & 1  & 0 & 0.3  & 0 \\
    $\theta_6\cup\theta_8$ & 0 & 0.75 & 0 & 0.25\\
    $\overline{\theta_4\cup\theta_8}$  & 0 & 0 & 0.7 & 0\\
    $I_t$ & 0 & 0.25 & 0 & 0.75\\
   \hline
  \end{tabular}
  \caption{Quantitative inputs of VBIED problem.}
\label{Table1}
\end{table}
 \begin{table}[!h]
 \small
\centering
 \begin{tabular}{|l|c|c|}
    \hline
    focal element     & $m_{PCR5}(.)$  & $m_{PCR6}(.)$ \\
    \hline  
    $\theta_1\cup\theta_2\cup\theta_3\cup\theta_5\cup\theta_6\cup\theta_7$ & 0.19741 &  0.16811\\
      $\theta_8$ & 0.24375 &  0.24375 \\
       $\theta_4\cup\theta_8$ & 0.33826 & 0.29641 \\    
       $\theta_6\cup\theta_8$ & 0.11029 &  0.14587\\    
           $I_t$ & 0.11029 & 0.14587 \\      
       \hline
  \end{tabular}
  \caption{Results of $m_0\oplus m_1\oplus m_2 \oplus m_3$ for Table \ref{Table1}.}
\label{ResultTable1}
\end{table}
    
\begin{table}[!h]
 \small
\centering
 \begin{tabular}{|l|c|c|}
    \hline
 Singletons       & $DSmP_{\epsilon,PCR5}(.)$  & $DSmP_{\epsilon,PCR6}(.)$ \\
    \hline 
$\theta_1$ &     0.0333   &  0.0286\\
$\theta_2$ &      0.0333  &  0.0286\\
$\theta_3$ &      0.0333  &  0.0286\\
$\theta_4$ &      0.0018  &  0.0018\\
$\theta_5$ &      0.0333  &  0.0286\\
$\theta_6$ &      0.0338  &  0.0292\\
$\theta_7$ &      0.0333  &  0.0286\\
$\theta_8$ &      0.7977  & 0.8260\\
       \hline
  \end{tabular}
  \caption{$DSmP_{\epsilon}$ of $m_0\oplus m_1\oplus m_2 \oplus m_3$ for Table \ref{Table1}.}
\label{ResultTable1DSmP}
\end{table}    

\begin{table}[!h]
 \small
\centering
 \begin{tabular}{|l|c|c|}
    \hline
 Singletons       & $BetP_{PCR5}(.)$  & $BetP_{PCR6}(.)$ \\
    \hline 
$\theta_1$ &     0.0467   &  0.0463\\
$\theta_2$ &      0.0467  & 0.0463\\
$\theta_3$ &      0.0467  &  0.0463\\
$\theta_4$ &      0.1829  &  0.1664\\
$\theta_5$ &     0.0467  &  0.0463\\
$\theta_6$ &      0.1018  & 0.1192\\
$\theta_7$ &      0.0467 &  0.0463\\
$\theta_8$ &      0.4818  & 0.4831\\
       \hline
  \end{tabular}
  \caption{$BetP$ of $m_0\oplus m_1\oplus m_2 \oplus m_3$ for Table \ref{Table1}.}
\label{ResultTable1BetP}
\end{table}

\medskip
\noindent
From fusion result of Table \ref{ResultTable1}, one gets for the danger supporting hypothesis $\theta_6\cup\theta_7\cup\theta_8$ (the worst scenario case)
\begin{itemize}
\item with PCR5: $\Delta(\theta_6\cup\theta_7\cup\theta_8)=0.64596$
$$P(\theta_6\cup\theta_7\cup\theta_8) \in [0.35404,1]$$
$$P(\overline{\theta_6\cup\theta_7\cup\theta_8}) \in [ 0,0.64596]$$
\item with PCR6:  $\Delta(\theta_6\cup\theta_7\cup\theta_8)=0.61038$
$$P(\theta_6\cup\theta_7\cup\theta_8)  \in [0.38962,1]$$
$$P(\overline{\theta_6\cup\theta_7\cup\theta_8})\in [0,0.61038]$$
\end{itemize}
\noindent
where $\Delta(X)=Pl(X)-Bel(X)$ is the imprecision related to $P(X)$. It is worth to note that $\Delta(\bar{X})=Pl(\bar{X})-Bel(\bar{X})=\Delta(X)$ because $Pl(\bar{X})=1-Bel(X)$ and $Bel(\bar{X})=1-Pl(X)$.\\

If we consider only $\theta_7 \cup \theta_8 = (\bar{A},V,B)\cup (A,V,B)$ as danger supporting hypothesis then from the fusion result of Table \ref{ResultTable1}, one gets
\begin{itemize}
\item with PCR5:  $ \Delta(\theta_7\cup\theta_8)=0.75625$
$$P(\theta_7\cup\theta_8) \in [ 0.24375,1]$$
$$P(\overline{\theta_7\cup\theta_8}) \in [ 0, 0.75625]$$
\item with PCR6:  $\Delta(\theta_7\cup\theta_8)=0.75625$
$$P(\theta_7\cup\theta_8) \in [ 0.24375,1]$$
$$P(\overline{\theta_7\cup\theta_8}) \in [ 0, 0.75625]$$
\end{itemize}

If we are more {\it{optimistic}} and we consider only the danger supporting hypothesis $\theta_8$, then one gets
\begin{itemize}
\item with PCR5: $\Delta(\theta_8)= 0.55884$
$$P(\theta_8) \in [0.24375, 0.80259]$$
$$P(\bar{\theta}_8) \in [0.19741,0.75625]$$
\item with PCR6:  $\Delta(\theta_8)=\Delta(\bar{\theta}_8)=0.58814$
$$P(\theta_8) \in [0.24375, 0.83189]$$
$$P(\bar{\theta}_8) \in [ 0.16811,0.75625]$$
\end{itemize}

If one approximates the bba's into probabilistic measures with DSmP transformation\footnote{DSmP transformation has been introduced and justified in details by the authors in the book \cite{DSmTBook1-3} (Vol.3, Chap. 3) freely downloadable from the web with many examples, and therefore it will not be presented here. }, one gets results with $\epsilon=0.001$ presented in Table \ref{ResultTable1DSmP}. One gets the higher probability on $\theta_8$ with respect to other alternatives and also $DSmP(\theta_6\cup\theta_7\cup\theta_8)=0.8648$. If one prefers to use the pignistic\footnote{BetP is the most used transformation to approximate a mass of belief into a subjective probability measure. It has been proposed by Philippe Smets in nineties.} probability transformation \cite{Smets1990}, one gets the results given in Table \ref{ResultTable1BetP}. One sees clearly that PIC\footnote{The PIC (probabilistic information content) criteria has been introduced by John Sudano in \cite{Sudano2002} and is noting but the dual of normalized Shannon entropy. $PIC$ is in $[0,1]$ and $PIC=1$ if the probability measure assigns a probability one only on a particular singleton of the frame, and $PIC=0$ if all elements of the frame are equi-probable.} of DSmP is higher that PIC of BetP which makes decision easier to take with DSmP than with BetP in favor of  $\theta_6\cup\theta_7\cup\theta_8$, or $\theta_7\cup\theta_8$, or $\theta_8$.

\begin{itemize}
\item[-] {\bf{Answer to Q1:}}
One sees that the result provided by PCR6 and PCR5 are very close and do not change fundamentally the final decision to take. Based on these very imprecise results, it is very difficult to take the right decision without decision-making error because the sources of information are highly uncertain and conflicting, but the analysis of lower and upper bounds shows that the most reasonable answer to the question based either on max of credibility or max of plausibility is to evacuate the building $B$ since $Bel(\theta_6\cup\theta_7\cup\theta_8)> Bel( \overline{\theta_6\cup\theta_7\cup\theta_8})$ and also $Pl(\theta_6\cup\theta_7\cup\theta_8)> Pl( \overline{\theta_6\cup\theta_7\cup\theta_8})$. The same conclusion is drawn when considering the element $\theta_7\cup\theta_8$ or $\theta_8$ alone. The same conclusion also is drawn (more easier) based on DSmP or on BetP values. In summary, the answer to Q1 is: {\bf{Evacuation of the building $B$}}.
\end{itemize}

In order to answer to the second question (Q2), let's compute the fusion results of the fusion $m_0\oplus m_2$ and  $m_0\oplus m_1\oplus m_3$ using inputs given in Table \ref{Table1}. The fusion results with corresponding DSmP and BetP are given in the Tables \ref{ResultExample1m0m2}-\ref{ResultExample1m0m2DSmP}.

\begin{table}[!h]
 \small
\centering
 \begin{tabular}{|l|c|c|}
    \hline
    focal element     & $m_{PCR5}(.)$  & $m_{PCR6}(.)$ \\
    \hline  
    $\theta_1\cup\theta_2\cup\theta_3\cup\theta_5\cup\theta_6\cup\theta_7$ & 0.28824 &  0.28824\\
       $\theta_4\cup\theta_8$ & 0.71176 & 0.71176 \\    
       \hline
  \end{tabular}
  \caption{Result of $m_0\oplus m_2$.}
\label{ResultExample1m0m2}
\end{table}

\begin{table}[!h]
 \small
\centering
 \begin{tabular}{|l|c|c|}
    \hline
 Singletons       & $DSmP_{\epsilon,PCR5}(.)$  & $DSmP_{\epsilon,PCR6}(.)$ \\
    \hline 
$\theta_1$ &      0.0480  &  0.0480\\
$\theta_2$ &      0.0480  &  0.0480\\
$\theta_3$ &      0.0480  &  0.0480\\
$\theta_4$ &      0.3560  &  0.3560\\
$\theta_5$ &      0.0480  &  0.0480\\
$\theta_6$ &      0.0480  &  0.0480\\
$\theta_7$ &      0.0480  &  0.0480\\
$\theta_8$ &      0.3560  &  0.3560\\
       \hline
  \end{tabular}
  \caption{$DSmP_{\epsilon}$ of $m_0\oplus m_2$.}
\label{ResultExample1m0m2DSmP}
\end{table}    

\begin{table}[!h]
 \small
\centering
 \begin{tabular}{|l|c|c|}
    \hline
 Singletons       & $BetP_{PCR5}(.)$  & $BetP_{PCR6}(.)$ \\
    \hline 
$\theta_1$ &      0.0480  &  0.0480\\
$\theta_2$ &      0.0480  &  0.0480\\
$\theta_3$ &      0.0480  &  0.0480\\
$\theta_4$ &      0.3560  &  0.3560\\
$\theta_5$ &      0.0480  &  0.0480\\
$\theta_6$ &      0.0480  &  0.0480\\
$\theta_7$ &      0.0480  &  0.0480\\
$\theta_8$ &      0.3560  &  0.3560\\
       \hline
  \end{tabular}
  \caption{$BetP$ of $m_0\oplus m_2$.}
\label{ResultExample1m0m2BetP}
\end{table}

Based on $m_0\oplus m_2$ fusion result, one gets a total imprecision $\Delta_{02}(\theta_6\cup\theta_7\cup\theta_8)=1$ when considering $\theta_6\cup\theta_7\cup\theta_8$ or $\theta_7\cup\theta_8$ since
$$P(\theta_6\cup\theta_7\cup\theta_8)\in [0,1]$$
$$P(\overline{\theta_6\cup\theta_7\cup\theta_8})\in [0, 1]$$
\noindent
and
$$P(\theta_7\cup\theta_8)\in [0,1]$$
$$P(\overline{\theta_7\cup\theta_8})\in [0, 1]$$
Even when considering only the danger supporting hypothesis $\theta_8$, one still gets a quite large imprecision on $P(\theta_8)$ since $\Delta_{02}(\theta_8)= 0.71176$ with
$$P(\theta_8)\in [0,0.71176]$$
$$P(\bar{\theta}_8)\in [0.28824, 1]$$
\noindent
Based on max of Bel or max of Pl criteria, one sees that it is not possible to take any rational decision from $\theta_6\cup\theta_7\cup\theta_8$ nor $\theta_7\cup\theta_8$ because of the full imprecision range of  $P(\theta_6\cup\theta_7\cup\theta_8)$ or $P(\theta_7\cup\theta_8)$. The decision using $m_0\oplus m_2$ (i.e. with prior information $m_0$ and ANPR system $m_2$) based only on supporting hypothesis $\theta_8$  should be to NOT evacuate the building $B$. Same decision would be taken based on DSmP or BetP values.
    
 \begin{table}[!h]
 \small
\centering
 \begin{tabular}{|l|c|c|}
    \hline
    focal element     & $m_{PCR5}(.)$  & $m_{PCR6}(.)$ \\
    \hline  
    $\theta_8$ & 0.8125 &  0.8125\\
       $\theta_4\cup\theta_8$ & 0.1875 & 0.1875 \\    
       \hline
  \end{tabular}
  \caption{Result of $m_0\oplus m_1\oplus m_3$.}
\label{ResultExample1m0m1m3}
\end{table}
   
\begin{table}[!h]
 \small
\centering
 \begin{tabular}{|l|c|c|}
    \hline
 Singletons       & $DSmP_{\epsilon,PCR5}(.)$  & $DSmP_{\epsilon,PCR6}(.)$ \\
    \hline 
$\theta_1$ &      0  &  0\\
$\theta_2$ &      0  &  0\\
$\theta_3$ &      0  &  0\\
$\theta_4$ &      0.0002  &  0.0002\\
$\theta_5$ &      0  &  0\\
$\theta_6$ &      0  &  0\\
$\theta_7$ &      0  &  0\\
$\theta_8$ &      0.9998  &  0.9998\\
       \hline
  \end{tabular}
  \caption{$DSmP_{\epsilon}$ of $m_0\oplus m_1\oplus m_3$.}
\label{ResultExample1m0m1m3DSmP}
\end{table}

\begin{table}[!h]
 \small
\centering
 \begin{tabular}{|l|c|c|}
    \hline
 Singletons       & $BetP_{PCR5}(.)$  & $BetP_{PCR6}(.)$ \\
    \hline 
$\theta_1$ &      0  &  0\\
$\theta_2$ &      0  &  0\\
$\theta_3$ &      0  &  0\\
$\theta_4$ &       0.09375  &   0.09375\\
$\theta_5$ &      0  &  0\\
$\theta_6$ &      0  &  0\\
$\theta_7$ &      0  &  0\\
$\theta_8$ &       0.90625  &   0.90625\\
       \hline
  \end{tabular}
  \caption{$BetP$ of $m_0\oplus m_1\oplus m_3$.}
\label{ResultExample1m0m1m3BetP}
\end{table}    

Based on $m_0\oplus m_1\oplus m_3$ fusion result, one gets $\Delta_{013}(\theta_6\cup\theta_7\cup\theta_8)=0.1875$
$$P(\theta_6\cup\theta_7\cup\theta_8)\in [0.8125,1]$$
$$P(\overline{\theta_6\cup\theta_7\cup\theta_8})\in [0, 0.1875]$$
\noindent
but also
$$P(\theta_7\cup\theta_8)\in [0.8125,1], \qquad P(\overline{\theta_7\cup\theta_8})\in [0, 0.1875]$$
\noindent
and 
$$P(\theta_8)\in [0.8125, 1], \qquad P(\bar{\theta}_8)\in [0, 0.1875]$$

Based on max of Bel or max of Pl criteria, the decision using $m_0\oplus m_1\oplus m_3$ (i.e. with prior information $m_0$ and both analysts) is to evacuate the building $B$. The same decision is taken based on DSmP or BetP values. It is worth to note that the precision on the result obtained with  $m_0\oplus m_1\oplus m_3$ is much better than with $m_0\oplus m_2$ since $\Delta_{013}(\theta_6\cup\theta_7\cup\theta_8) < \Delta_{02}(\theta_6\cup\theta_7\cup\theta_8)$, or $\Delta_{013}(\theta_7\cup\theta_8) < \Delta_{02}(\theta_7\cup\theta_8)$. Moreover it is easy to verify that $m_0\oplus m_1\oplus m_3$ fusion system is more informative than $m_0\oplus m_2$ fusion system because Shannon entropy of DSmP of $m_0\oplus m_2$ is much bigger than Shannon entropy of DSmP of $m_0\oplus m_1\oplus m_3$.
\begin{itemize}
\item[-]{\bf{Answer to Q2:}}  Since the information obtained by the fusion $m_0\oplus m_2$ is less informative and less precise than the information obtained with the fusion $m_0\oplus m_1\oplus m_3$, it is better to choose and to trust the fusion system $m_0\oplus m_1\oplus m_3$ rather than $m_0\oplus m_2$. Based on this choice, the final decision will be to evacuate the building $B$ which is consistent with answer to question Q1.\\
\end{itemize}

\noindent
{\bf{Example 2}}: Let's modify a bit the previous Table \ref{Table1} and take higher belief for sources 1 and 3 as
 \begin{table}[!h]
 \small
\centering
 \begin{tabular}{|l|c|c|c|c|}
    \hline
    focal element     & $m_{0}(.)$  & $m_{1}(.)$ & $m_{2}(.)$ & $m_{3}(.)$ \\
    \hline  
    $\theta_4\cup\theta_8$  & 1  & 0 & 0.3  & 0 \\
    $\theta_6\cup\theta_8$ & 0 & 0.9 & 0 & 0.1\\
    $\overline{\theta_4\cup\theta_8}$  & 0 & 0 & 0.7 & 0\\
    $I_t$ & 0 & 0.1 & 0 & 0.9\\
   \hline
  \end{tabular}
  \caption{Quantitative inputs of VBIED problem.}
\label{Table2}
\end{table}

\noindent
The results of the fusion $m_0\oplus m_1\oplus m_2 \oplus m_3$ using PCR5 and PCR6 and the corresponding DSmP values are given in tables \ref{ResultTable2}-\ref{ResultTable2DSmP}.

 \begin{table}[!h]
 \small
\centering
 \begin{tabular}{|l|c|c|}
    \hline
    focal element     & $m_{PCR5}(.)$  & $m_{PCR6}(.)$ \\
    \hline  
       $\theta_1\cup\theta_2\cup\theta_3\cup\theta_5\cup\theta_6\cup\theta_7$ & 0.16525 &  0.14865\\
       $\theta_8$ & 0.27300 & 0.27300 \\
       $\theta_4\cup\theta_8$ & 0.26307 & 0.23935 \\    
       $\theta_6\cup\theta_8$ & 0.14934 & 0.16950\\    
       $I_t$ & 0.14934 & 0.16950 \\      
       \hline
  \end{tabular}
  \caption{Results of $m_0\oplus m_1\oplus m_2 \oplus m_3$ for Table \ref{Table2}.}
\label{ResultTable2}
\end{table}
    
 \begin{table}[!h]
 \small
\centering
 \begin{tabular}{|l|c|c|}
    \hline
 Singletons       & $DSmP_{\epsilon,PCR5}(.)$  & $DSmP_{\epsilon,PCR6}(.)$ \\
    \hline 
$\theta_1$ &      0.0281  &  0.0254\\
$\theta_2$ &      0.0281  & 0.0254\\
$\theta_3$ &      0.0281  & 0.0254\\
$\theta_4$ &      0.0015  & 0.0015\\
$\theta_5$ &      0.0281  &  0.0254\\
$\theta_6$ &      0.0286  &  0.0260\\
$\theta_7$ &      0.0281  &  0.0254\\
$\theta_8$ &      0.8294  &  0.8455\\
       \hline
  \end{tabular}
  \caption{$DSmP_{\epsilon}$ of $m_0\oplus m_1\oplus m_2 \oplus m_3$ for Table \ref{Table2}.}
\label{ResultTable2DSmP}
\end{table}

\begin{table}[!h]
 \small
\centering
 \begin{tabular}{|l|c|c|}
    \hline
 Singletons       & $BetP_{PCR5}(.)$  & $BetP_{PCR6}(.)$ \\
    \hline 
$\theta_1$ &     0.0462   &  0.0460\\
$\theta_2$ &      0.0462  & 0.0460\\
$\theta_3$ &      0.0462  &  0.0460\\
$\theta_4$ &      0.1502  &  0.1409\\
$\theta_5$ &     0.0462  &  0.0460\\
$\theta_6$ &      0.1209  & 0.1307\\
$\theta_7$ &      0.0462 &  0.0460\\
$\theta_8$ &      0.4979  & 0.4986\\
       \hline
  \end{tabular}
  \caption{$BetP$ of $m_0\oplus m_1\oplus m_2 \oplus m_3$ for Table \ref{Table2}.}
\label{ResultTable2BetP}
\end{table}    

\noindent
From fusion result of Table \ref{ResultTable2}, one gets when considering $\theta_6\cup\theta_7\cup \theta_8$
\begin{itemize}
\item with PCR5:  $ \Delta(\theta_6\cup\theta_7\cup\theta_8)=0.57766$
$$P(\theta_6\cup\theta_7\cup\theta_8) \in [ 0.42234,1]$$
$$P(\overline{\theta_6\cup\theta_7\cup\theta_8}) \in [ 0,57766]$$
\item with PCR6:  $\Delta(\theta_7\cup\theta_8)=0.5575$
$$P(\theta_6\cup\theta_7\cup\theta_8) \in [ 0.4425,1]$$
$$P(\overline{\theta_6\cup\theta_7\cup\theta_8}) \in [ 0,0.5575]$$
\end{itemize}
\noindent
and when considering $\theta_7\cup \theta_8$ 
\begin{itemize}
\item with PCR5:  $ \Delta(\theta_7\cup\theta_8)=0.7270$
$$P(\theta_7\cup\theta_8) \in [ 0.27300,1], \quad P(\overline{\theta_7\cup\theta_8}) \in [ 0,0.7270]$$
\item with PCR6:  $\Delta(\theta_7\cup\theta_8)=0.7270$
$$P(\theta_7\cup\theta_8) \in [ 0.27300,1], \quad P(\overline{\theta_7\cup\theta_8}) \in [ 0,0.7270]$$
\end{itemize}
\noindent
and when considering $\theta_8$ only
\begin{itemize}
\item with PCR5 or PCR6: $\Delta(\theta_8)=0.56175$
$$P(\theta_8) \in [0.27300, 0.83475]$$
$$P(\bar{\theta}_8) \in [ 0.16525,0.7270]$$
\item with PCR6:  $\Delta(\theta_8)=0.57835$
$$P(\theta_8) \in [ 0.27300,0.85135]$$
$$P(\bar{\theta}_8) \in [0.14865,0.7270]$$
\end{itemize}

\noindent
One gets also the following $DSmP_{\epsilon=0.001}$ values
\begin{align*}
DSmP_{\epsilon,PCR5}(\theta_6\cup\theta_7\cup\theta_8)&= 0.8861\\
DSmP_{\epsilon,PCR6}(\theta_6\cup\theta_7\cup\theta_8)&=0.8869\\
DSmP_{\epsilon,PCR5}(\theta_7\cup\theta_8)&= 0.8575\\
DSmP_{\epsilon,PCR6}(\theta_7\cup\theta_8)&=0.8709\\
DSmP_{\epsilon,PCR5}(\theta_8)&= 0.8294\\
DSmP_{\epsilon,PCR6}(\theta_8)&= 0.8455
\end{align*}

\begin{itemize}
\item[-] {\bf{Answer to Q1:}}
Using an analysis similar to the one done for Example 1, based on max of credibility or max of plausibility criteria, or by considering the DSmP or BetP values of $\theta_6\cup\theta_7\cup\theta_8$, or $\theta_7\cup\theta_8$, or $\theta_8$ the decision to take is: {\bf{Evacuate the building $B$}}.
\end{itemize}
    
In order to answer to the second question (Q2) for this Example 2, let's compute the fusion results of the fusion $m_0\oplus m_2$ and  $m_0\oplus m_1\oplus m_3$ using inputs given in Table \ref{Table2}. Since the inputs $m_0$ and $m_2$ are the same as those in Example 1, the $m_0\oplus m_2$ fusion results with corresponding DSmP are those already given in Tables \ref{ResultExample1m0m2}-\ref{ResultExample1m0m2BetP}. Only the fusion  $m_0\oplus m_1\oplus m_3$ must be derived with the new bba's $m_1$ and $m_3$ chosen for this Example 2.
The $m_0\oplus m_1\oplus m_3$ fusion results obtained with PCR5 and PCR6, and the corresponding DSmP and BetP values are shown in Tables \ref{ResultExample2m0m1m3}-\ref{ResultExample2m0m1m3BetP}. According to these results, one gets with the PCR5 or PCR6 fusion $m_0\oplus m_1\oplus m_3$: $\Delta_{013}(\theta_6\cup\theta_7\cup\theta_8)=\Delta_{013}(\theta_7\cup\theta_8)=\Delta_{013}(\theta_8)=0.09$ and 
$$P(\theta_6\cup\theta_7\cup\theta_8)\in [0.91,1], \quad P(\overline{\theta_6\cup\theta_7\cup\theta_8})\in [0, 0.09]$$
$$P(\theta_7\cup\theta_8)\in [0.91,1], \quad P(\overline{\theta_7\cup\theta_8})\in [0, 0.09]$$
$$P(\theta_8)\in [0.91, 1], \quad P(\bar{\theta}_8)\in [0, 0.09]$$

 \begin{table}[!h]
 \small
\centering
 \begin{tabular}{|l|c|c|}
    \hline
    focal element     & $m_{PCR5}(.)$  & $m_{PCR6}(.)$ \\
    \hline  
    $\theta_8$ & 0.91 &  0.91\\
       $\theta_4\cup\theta_8$ & 0.09 & 0.09 \\    
       \hline
  \end{tabular}
  \caption{Result of $m_0\oplus m_1\oplus m_3$.}
\label{ResultExample2m0m1m3}
\end{table}
   
\begin{table}[!h]
 \small
\centering
 \begin{tabular}{|l|c|c|}
    \hline
 Singletons       & $DSmP_{\epsilon,PCR5}(.)$  & $DSmP_{\epsilon,PCR6}(.)$ \\
    \hline 
$\theta_1$ &      0  &  0\\
$\theta_2$ &      0  &  0\\
$\theta_3$ &      0  &  0\\
$\theta_4$ &      0.0001  &  0.0001\\
$\theta_5$ &      0  &  0\\
$\theta_6$ &      0  &  0\\
$\theta_7$ &      0  &  0\\
$\theta_8$ &      0.9999  &  0.9999\\
       \hline
  \end{tabular}
  \caption{$DSmP_{\epsilon}$ of $m_0\oplus m_1\oplus m_3$.}
\label{ResultExample2m0m1m3DSmP}
\end{table}

\begin{table}[!h]
 \small
\centering
 \begin{tabular}{|l|c|c|}
    \hline
 Singletons       & $BetP_{PCR5}(.)$  & $BetP_{PCR6}(.)$ \\
    \hline 
$\theta_1$ &      0  &  0\\
$\theta_2$ &      0  &  0\\
$\theta_3$ &      0  &  0\\
$\theta_4$ &      0.045  &  0.045\\
$\theta_5$ &      0  &  0\\
$\theta_6$ &      0  &  0\\
$\theta_7$ &      0  &  0\\
$\theta_8$ &      0.955  &  0.955\\
       \hline
  \end{tabular}
  \caption{$BetP$ of  $m_0\oplus m_1\oplus m_3$.}
\label{ResultExample2m0m1m3BetP}
\end{table}

Based on max of Bel or max of Pl criteria, the decision using $m_0\oplus m_1\oplus m_3$ (i.e. with prior information $m_0$ and both analysts) is to evacuate the building $B$. Same decision is taken based on DSmP or BetP values. It is worth to note that the precision on the result obtained with  $m_0\oplus m_1\oplus m_3$ is much better than with $m_0\oplus m_2$ since $\Delta_{013}(\theta_8) < \Delta_{02}(\theta_8)$, or $\Delta_{013}(\theta_7\cup\theta_8) < \Delta_{02}(\theta_7\cup\theta_8)$, or $\Delta_{013}(\theta_6\cup\theta_7\cup\theta_8) < \Delta_{02}(\theta_6\cup\theta_7\cup\theta_8)$. Moreover it is easy to verify that $m_0\oplus m_1\oplus m_3$ fusion system is more informative than $m_0\oplus m_2$ fusion system because Shannon entropy of DSmP of $m_0\oplus m_2$ is much bigger than Shannon entropy of DSmP of $m_0\oplus m_1\oplus m_3$. Same remark holds with BetP transformation.

\begin{itemize}
\item[-]{\bf{Answer to Q2:}}  Since the information obtained by the fusion $m_0\oplus m_2$ is less informative and less precise than the information obtained with the fusion $m_0\oplus m_1\oplus m_3$, it is better to choose and to trust the fusion system $m_0\oplus m_1\oplus m_3$ rather than $m_0\oplus m_2$. Based on this choice, the final decision will be to evacuate the building $B$ which is consistent with the answer of the question Q1.
\end{itemize}
   
\subsection{Impact of prior information}

To see the impact of the quality/reliability of prior information on the result, let's modify the input $m_0(.)$ in previous Tables \ref{Table1} and \ref{Table2} and consider now a very uncertain prior source.\\


\noindent
{\bf{Example 3}}: We consider the very uncertain prior source of information $m_0( \theta_{4}\cup \theta_{8})=0.1$ and $m_0(I_t)=0.9$. The results for the modified inputs Table \ref{Table3} ate given in Tables \ref{ResultTable3} and \ref{ResultTable3DSmP}.\\

 \begin{table}[!h]
 \small
\centering
 \begin{tabular}{|l|c|c|c|c|}
    \hline
    focal element     & $m_{0}(.)$  & $m_{1}(.)$ & $m_{2}(.)$ & $m_{3}(.)$ \\
    \hline  
    $\theta_4\cup\theta_8$  & 0.1  & 0 & 0.3  & 0 \\
    $\theta_6\cup\theta_8$ & 0 & 0.75 & 0 & 0.25\\
    $\overline{\theta_4\cup\theta_8}$  & 0 & 0 & 0.7 & 0\\
    $I_t$ & 0.9 & 0.25 & 0 & 0.75\\
   \hline
  \end{tabular}
  \caption{Quantitative inputs of VBIED problem.}
\label{Table3}
\end{table}

 \begin{table}[!h]
 \small
\centering
 \begin{tabular}{|l|c|c|}
    \hline
    focal element     & $m_{PCR5}(.)$  & $m_{PCR6}(.)$ \\
    \hline  
     $\theta_6$ & 0.511870 &  0.511870\\
     $\theta_1\cup\theta_2\cup\theta_3\cup\theta_5\cup\theta_6\cup\theta_7$ & 0.151070 &  0.142670\\
      $\theta_8$ & 0.243750 & 0.243750 \\
       $\theta_4\cup\theta_8$ & 0.060957 & 0.059757 \\    
       $\theta_6\cup\theta_8$ & 0.016173 & 0.020973\\    
           $I_t$ & 0.016173 & 0.020973 \\      
       \hline
  \end{tabular}
  \caption{Result of $m_0\oplus m_1\oplus m_2\oplus m_3$ for Table \ref{Table3}.}
\label{ResultTable3}
\end{table}    
    
 \begin{table}[!h]
 \small
\centering
 \begin{tabular}{|l|c|c|}
    \hline
 Singletons       & $DSmP_{\epsilon,PCR5}(.)$  & $DSmP_{\epsilon,PCR6}(.)$ \\
    \hline 
$\theta_1$ &      0.0003   & 0.0003\\
$\theta_2$ &      0.0003   & 0.0003\\
$\theta_3$ &      0.0003   & 0.0003\\
$\theta_4$ &      0.0003   & 0.0003\\
$\theta_5$ &      0.0003   & 0.0003\\
$\theta_6$ &      0.6833   & 0.6815\\
$\theta_7$ &      0.0003   &  0.0003\\
$\theta_8$ &      0.3149   &  0.3168\\
       \hline
  \end{tabular}
 \caption{$DSmP_{\epsilon}$ of $m_0\oplus m_1\oplus m_2\oplus m_3$ for Table \ref{Table3}.}
\label{ResultTable3DSmP}
\end{table}

         \begin{table}[!h]
 \small
\centering
 \begin{tabular}{|l|c|c|}
    \hline
 Singletons       & $BetP_{PCR5}(.)$  & $BetP_{PCR6}(.)$ \\
    \hline 
$\theta_1$ &     0.0272   & 0.0264\\
$\theta_2$ &      0.0272   & 0.0264\\
$\theta_3$ &      0.0272   & 0.0264\\
$\theta_4$ &     0.0325  & 0.0325\\
$\theta_5$ &     0.0272   & 0.0264\\
$\theta_6$ &      0.5472  & 0.5488\\
$\theta_7$ &      0.0272 &  0.0264\\
$\theta_8$ &      0.2843   &  0.2867\\
       \hline
  \end{tabular}
 \caption{$BetP$ of $m_0\oplus m_1\oplus m_2\oplus m_3$ for Table \ref{Table3}.}
\label{ResultTable3BetP}
\end{table}       

\noindent
From the fusion result of Table \ref{ResultTable3}, one gets when considering $\theta_6\cup\theta_7\cup \theta_8$ 
\begin{itemize}
\item with PCR5: $\Delta(\theta_6\cup\theta_7\cup\theta_8)=0.221377$
$$P(\theta_6\cup\theta_7\cup\theta_8) \in [ 0.778623, 1]$$
$$P(\overline{\theta_6\cup\theta_7\cup\theta_8}) \in [ 0,0.221377]$$
\item with PCR6:  $\Delta(\theta_6\cup\theta_7\cup\theta_8)=0.21465$
$$P(\theta_6\cup\theta_7\cup\theta_8) \in [ 0.78535,1] $$
$$P(\overline{\theta_6\cup\theta_7\cup\theta_8}) \in [ 0,0.21465]$$
\end{itemize}
\noindent
and when considering $\theta_7\cup \theta_8$ 
\begin{itemize}
\item with PCR5: $\Delta(\theta_7\cup\theta_8)=0.24438$
$$P(\theta_7\cup\theta_8) \in [ 0.24375, 0.48813]$$
$$P(\overline{\theta_7\cup\theta_8}) \in [ 0.51187,0.75625]$$
\item with PCR6: $\Delta(\theta_7\cup\theta_8)=0.24438$
$$P(\theta_7\cup\theta_8) \in [ 0.24375, 0.48813]$$
$$P(\overline{\theta_7\cup\theta_8}) \in [ 0.51187,0.75625]$$
\end{itemize}
\noindent
and when considering $\theta_8$ only, one has
\begin{itemize}
\item with PCR5: $\Delta(\theta_8)=0.09331$
$$P(\theta_8) \in [0.24375, 0.33706]$$
$$P(\bar{\theta}_8) \in [ 0.66294,0.75625]$$
\item with PCR6:  $\Delta(\theta_8)=0.10171$
$$P(\theta_8) \in [  0.24375, 0.34546 ]$$
$$P(\bar{\theta}_8) \in [0.65454,0.75625]$$
\end{itemize}

\noindent
Using DSmP transformation, one gets a low probability in $ \theta_8$ or in $\theta_7\cup\theta_8$ because
\begin{align*}
DSmP_{\epsilon,PCR5}(\theta_6\cup\theta_7\cup\theta_8)&= 0.9985\\
DSmP_{\epsilon,PCR6}(\theta_6\cup\theta_7\cup\theta_8)&=0.9986\\
DSmP_{\epsilon,PCR5}(\theta_7\cup\theta_8)&= 0.3152\\
DSmP_{\epsilon,PCR6}(\theta_7\cup\theta_8)&=0.3171\\
DSmP_{\epsilon,PCR5}(\theta_8)&= 0.3149\\
DSmP_{\epsilon,PCR6}(\theta_8)&= 0.3168
\end{align*}

\begin{itemize}
\item[-] {\bf{Answer to Q1:}} The analysis of these results are very interesting since one sees that the element of the frame $\Theta$ having the highest DSmP (or BetP)  is $\theta_6=(A,\bar{V},B)$ and it has a very strong impact on the final decision. Because if one considers only $ \theta_8$ or  $\theta_7\cup\theta_8$ has decision-support hypotheses, one sees that the decision to take is to NOT evacuate the building $B$ since one gets a low probability in $ \theta_8$ or in $\theta_7\cup\theta_8$. Whereas if we include also $ \theta_6$ in the decision-support hypothesis, then the final decision will be the opposite since $DSmP(\theta_6\cup\theta_7\cup\theta_8)$ is very close to one with PCR5 or with PCR6. The same behavior occurs with BetP. So there is a strong impact of prior information on the final decision since without strong prior information supporting $\theta_4\cup\theta_8$ we have to conclude either to the non evacuation of building $B$ based on the max of credibility, the max of plausibility or the max of DSmP using $\theta_8$ or $\theta_7\cup\theta_8$ for decision-making, or to the evacuation of the building if a more prudent strategy is used based on $\theta_6\cup\theta_7\cup\theta_8$ decision-support hypothesis.
\end{itemize}

Let's examine the results of fusion systems $m_0\oplus m_2$ and $m_0\oplus m_1\oplus m_3$ given in Tables \ref{ResultExample3m0m2}-\ref{ResultExample3m0m1m3BetP}.

 \begin{table}[!h]
 \small
\centering
 \begin{tabular}{|l|c|c|}
    \hline
    focal element     & $m_{PCR5}(.)$  & $m_{PCR6}(.)$ \\
    \hline  
      $\theta_1\cup\theta_2\cup\theta_3\cup\theta_5\cup\theta_6\cup\theta_7$ & 0.69125 &  0.69125\\
       $\theta_4\cup\theta_8$ & 0.30875 & 0.30875 \\    
       \hline
  \end{tabular}
  \caption{Result of $m_0\oplus m_2$.}
\label{ResultExample3m0m2}
\end{table}

\begin{table}[!h]
 \small
\centering
 \begin{tabular}{|l|c|c|}
    \hline
 Singletons       & $DSmP_{\epsilon,PCR5}(.)$  & $DSmP_{\epsilon,PCR6}(.)$ \\
    \hline 
$\theta_1$ &      0.1152  &   0.1152\\
$\theta_2$ &      0.1152 &   0.1152\\
$\theta_3$ &      0.1152 &   0.1152\\
$\theta_4$ &      0.1544 &   0.1544\\
$\theta_5$ &      0.1152 &   0.1152\\
$\theta_6$ &      0.1152 &   0.1152\\
$\theta_7$ &      0.1152  &  0.1152\\
$\theta_8$ &      0.1544  &   0.1544\\
       \hline
  \end{tabular}
  \caption{$DSmP_{\epsilon}$ of $m_0\oplus m_2$.}
\label{ResultExample3m0m2DSmP}
\end{table}  

\begin{table}[!h]
 \small
\centering
 \begin{tabular}{|l|c|c|}
    \hline
 Singletons       & $BetP_{\epsilon,PCR5}(.)$  & $BetP_{\epsilon,PCR6}(.)$ \\
    \hline 
$\theta_1$ &      0.1152  &   0.1152\\
$\theta_2$ &      0.1152 &   0.1152\\
$\theta_3$ &      0.1152 &   0.1152\\
$\theta_4$ &      0.1544 &   0.1544\\
$\theta_5$ &      0.1152 &   0.1152\\
$\theta_6$ &      0.1152 &   0.1152\\
$\theta_7$ &      0.1152  &  0.1152\\
$\theta_8$ &      0.1544  &   0.1544\\
       \hline
  \end{tabular}
  \caption{$BetP$ of $m_0\oplus m_2$.}
\label{ResultExample3m0m2BetP}
\end{table}  
   
 \begin{table}[!h]
 \small
\centering
 \begin{tabular}{|l|c|c|}
    \hline
    focal element     & $m_{PCR5}(.)$  & $m_{PCR6}(.)$ \\
    \hline  
       $\theta_8$ & 0.08125 &  0.08125\\
       $\theta_4\cup\theta_8$ & 0.01875 & 0.01875 \\    
       $\theta_6\cup\theta_8$ & 0.73125 & 0.73125 \\    
        $I_t$ & 0.16875 & 0.16875 \\    
     \hline
  \end{tabular}
  \caption{Result of $m_0\oplus m_1\oplus m_3$.}
\label{ResultExample3m0m1m3}
\end{table}

\begin{table}[!h]
 \small
\centering
 \begin{tabular}{|l|c|c|}
    \hline
 Singletons       & $DSmP_{\epsilon,PCR5}(.)$  & $DSmP_{\epsilon,PCR6}(.)$ \\
    \hline 
$\theta_1$ &      0.0019 &   0.0019\\
$\theta_2$ &      0.0019 &   0.0019\\
$\theta_3$ &      0.0019  &  0.0019\\
$\theta_4$ &      0.0021  &  0.0021\\
$\theta_5$ &      0.0019 &   0.0019\\
$\theta_6$ &      0.0107  &  0.0107\\
$\theta_7$ &      0.0019 &   0.0019\\
$\theta_8$ &      0.9778  &  0.9778\\
       \hline
  \end{tabular}
  \caption{$DSmP_{\epsilon}$ of $m_0\oplus m_1\oplus m_3$.}
\label{ResultExample3m0m1m3DSmP}
\end{table}    

\begin{table}[!h]
 \small
\centering
 \begin{tabular}{|l|c|c|}
    \hline
 Singletons       & $BetP_{\epsilon,PCR5}(.)$  & $BetP_{\epsilon,PCR6}(.)$ \\
    \hline 
$\theta_1$ &     0.0211  &   0.0211\\
$\theta_2$ &      0.0211  &   0.0211\\
$\theta_3$ &      0.0211  &   0.0211\\
$\theta_4$ &      0.0305 &   0.0305\\
$\theta_5$ &      0.0211  &   0.0211\\
$\theta_6$ &     0.3867 &  0.3867\\
$\theta_7$ &      0.0211  &   0.0211\\
$\theta_8$ &      0.4773  &   0.4773\\
       \hline
  \end{tabular}
  \caption{$BetP$ of $m_0\oplus m_1\oplus m_3$.}
\label{ResultExample3m0m1m3BetP}
\end{table}

Based on $m_0\oplus m_2$ fusion result, one gets a large imprecision on evaluation of probabilities of decision-support hypotheses since for $\theta_6\cup\theta_7\cup\theta_8$, one has
$\Delta_{02}(\theta_6\cup\theta_7\cup\theta_8)=1$ and
$$P(\theta_6\cup\theta_7\cup\theta_8)\in [0,1]$$
$$P(\overline{\theta_6\cup\theta_7\cup\theta_8})\in [0, 1]$$
\noindent
for $\theta_7\cup\theta_8$, one has also $\Delta_{02}(\theta_7\cup\theta_8)=1$ with
$$P(\theta_7\cup\theta_8)\in [0,1]$$
$$P(\overline{\theta_7\cup\theta_8})\in [0, 1]$$
\noindent
and for $\theta_8$, one gets $\Delta_{02}(\theta_8)=  0.30875$ with
$$P(\theta_8)\in [0, 0.30875]$$
$$P(\bar{\theta}_8)\in [0.69125, 1]$$

One sees that it is impossible to take a decision when considering only $\theta_6\cup\theta_7\cup\theta_8$ or $\theta_7\cup\theta_8$ because of full imprecision of the corresponding probabilities.
However, based on max of Bel or max of Pl criteria on $\theta_8$ the decision using $m_0\oplus m_2$ (i.e. with uncertain prior information $m_0$ and ANPR system $m_2$) is to NOT evacuate the building $B$. According to Tables \ref{ResultExample3m0m2DSmP}-\ref{ResultExample3m0m2BetP}, one sees also an ambiguity in decision-making between $\theta_8$ and $\theta_4$ since they have the same DSmP (or BetP) values. \\

Based on $m_0\oplus m_1\oplus m_3$ fusion results given in Tables \ref{ResultExample3m0m1m3}-\ref{ResultExample3m0m1m3BetP}, one gets for $\theta_6\cup\theta_7\cup\theta_8$ the imprecision $\Delta_{013}(\theta_6\cup\theta_7\cup\theta_8)=0.1875$ with
$$P(\theta_6\cup\theta_7\cup\theta_8)\in [0.8125,1]$$
$$P(\overline{\theta_6\cup\theta_7\cup\theta_8})\in [0, 0.1875]$$
\noindent
for $\theta_7\cup\theta_8$, one gets $\Delta_{013}(\theta_7\cup\theta_8)=0.91875$ with
$$P(\theta_7\cup\theta_8)\in [0.08125, 1]$$
$$P(\overline{\theta_7\cup\theta_8})\in [0,0.91875]$$
\noindent
and for $\theta_8$, one gets $\Delta_{013}(\theta_8)=  0.91875$ with
$$P(\theta_8)\in [0.08125, 1]$$
$$P(\bar{\theta}_8)\in [0,0.91875]$$

Based on max of Bel or max of Pl criteria, the decision using $m_0\oplus m_1\oplus m_3$ is the evacuation of the building $B$. Same decision is drawn when using DSmP or BetP results according to Tables \ref{ResultExample3m0m1m3DSmP} and \ref{ResultExample3m0m1m3BetP}. With this uncertain prior information, it is worth to note that the precision on the result obtained with  $m_0\oplus m_1\oplus m_3$ is better than with $m_0\oplus m_2$ when considering (in cautious strategy) the decision-support hypotheses $ \theta_6\cup\theta_7\cup\theta_8$ or $ \theta_7\cup\theta_8$ since $\Delta_{013}( \theta_6\cup\theta_7\cup\theta_8) < \Delta_{02}( \theta_6\cup\theta_7\cup\theta_8)$, or $\Delta_{013}(\theta_7\cup\theta_8) < \Delta_{02}(\theta_7\cup\theta_8)$. 
However, if a more optimistic/risky strategy is used when considering only $\theta_8$ as decision-support hypothesis, it is preferable to choose the subsystem  $m_0\oplus m_2$ because $\Delta_{02}(\theta_8) < \Delta_{013}(\theta_8)$. However, one sees that globally $m_0\oplus m_1\oplus m_3$ fusion system is more informative than $m_0\oplus m_2$ fusion system because Shannon entropy of DSmP of $m_0\oplus m_2$ is much bigger than Shannon entropy of DSmP of $m_0\oplus m_1\oplus m_3$.

\begin{itemize}
\item[-]{\bf{Answer to Q2:}} The answer of question Q2 is not easy because it depends both on the criterion (precision or PIC) and on the decision-support hypothesis we choose. Based on precision criterion and taking the optimistic point of view using only $\theta_8$, it is better to trust $m_0\oplus m_2$ fusion system since $\Delta_{02}(\theta_8)=\Delta_{02}(\bar{\theta}_8)=  0.30875$ whereas $\Delta_{013}(\theta_8)=\Delta_{013}(\bar{\theta}_8)= 0.9187$. In such case, one should NOT evacuate the building $B$. If we consider that is better to trust result of $m_0\oplus m_1\oplus m_3$ fusion system because it is more informative than $m_0\oplus m_2$ then the decision should be to evacuate the building $B$. If we take a more prudent point of view in considering as decision-support hypotheses  either $ \theta_6\cup\theta_7\cup\theta_8$ or $ \theta_7\cup\theta_8$, then the final decision taken according to the (most precise and informative) subsystem $m_0\oplus m_1\oplus m_3$ is to evacuate the building $B$. 

So the main open question is what solution to choose for selecting either $m_0\oplus m_2$ or $m_0\oplus m_1\oplus m_3$  fusion system ? In authors opinion, in such case it seems better to base our choice on the precision level of information one has really in hands (rather than the PIC value which is always related to some ad-hoc probabilistic transformation) and in adopting the most prudent strategy. Therefore for this example, the final decision must be done according to $m_0\oplus m_1\oplus m_3$, i.e. evacuate the building $B$.
\end{itemize}


\noindent
{\bf{Example 4}}:  Let's modify a bit the previous input Table \ref{Table3} and take higher belief for sources 1 and 3 as
 \begin{table}[!h]
 \small
\centering
 \begin{tabular}{|l|c|c|c|c|}
    \hline
    focal element     & $m_{0}(.)$  & $m_{1}(.)$ & $m_{2}(.)$ & $m_{3}(.)$ \\
    \hline  
    $\theta_4\cup\theta_8$        & 0.1  & 0    & 0.3  & 0 \\
    $\theta_6\cup\theta_8$       & 0     & 0.9 & 0     & 0.1\\
    $\overline{\theta_4\cup\theta_8}$        & 0     & 0    & 0.7  & 0\\
    $I_t$ & 0.9 & 0.1 & 0    & 0.9\\
   \hline
  \end{tabular}
  \caption{Quantitative inputs of VBIED problem.}
\label{Table4}
\end{table}

 \begin{table}[!h]
 \small
\centering
 \begin{tabular}{|l|c|c|}
    \hline
    focal element     & $m_{PCR5}(.)$  & $m_{PCR6}(.)$ \\
    \hline  
     $\theta_6$ & 0.573300 &  0.573300\\
    $\theta_1\cup\theta_2\cup\theta_3\cup\theta_5\cup\theta_6\cup\theta_7$ & 0.082365 &  0.077355\\
      $\theta_8$ & 0.273000 & 0.273000 \\
       $\theta_4\cup\theta_8$ & 0.030666 & 0.029951 \\    
       $\theta_6\cup\theta_8$ & 0.020334 & 0.023197\\    
           $I_t$ & 0.020334 & 0.023197 \\      
       \hline
  \end{tabular}
  \caption{Result of $m_0\oplus m_1\oplus m_2 \oplus m_3$ for Table \ref{Table4}.}
\label{ResultTable4}
\end{table}    
    
 \begin{table}[!h]
 \small
\centering
 \begin{tabular}{|l|c|c|}
    \hline
 Singletons       & $DSmP_{\epsilon,PCR5}(.)$  & $DSmP_{\epsilon,PCR6}(.)$ \\
    \hline 
$\theta_1$ &      0.0002   & 0.0002\\
$\theta_2$ &      0.0002   & 0.0002\\
$\theta_3$ &      0.0002   & 0.0002\\
$\theta_4$ &      0.0001   & 0.0001\\
$\theta_5$ &      0.0002   & 0.0002\\
$\theta_6$ &      0.6824   & 0.6813\\
$\theta_7$ &      0.0002   & 0.0002\\
$\theta_8$ &      0.3166   & 0.3178\\
       \hline
  \end{tabular}
  \caption{$DSmP_{\epsilon}$ of $m_0\oplus m_1\oplus m_2\oplus m_3$ for Table \ref{Table4}.}
\label{ResultTable4DSmP}
\end{table}       

 \begin{table}[!h]
 \small
\centering
 \begin{tabular}{|l|c|c|}
    \hline
 Singletons       & $BetP_{PCR5}(.)$  & $BetP_{PCR6}(.)$ \\
    \hline 
$\theta_1$ &      0.0163 &   0.0158\\
$\theta_2$ &      0.0163 &   0.0158\\
$\theta_3$ &      0.0163 &   0.0158\\
$\theta_4$ &      0.0179 &   0.0179\\
$\theta_5$ &      0.0163 &   0.0158\\
$\theta_6$ &      0.5997 &   0.6007\\
$\theta_7$ &      0.0163 &    0.0158\\
$\theta_8$ &      0.3010 &   0.3025\\
       \hline
  \end{tabular}
  \caption{$BetP$ of $m_0\oplus m_1\oplus m_2\oplus m_3$ for Table \ref{Table4}.}
\label{ResultTable4BetP}
\end{table}       

\noindent
Therefore, one gets when considering $\theta_6\cup\theta_7\cup \theta_8$
\begin{itemize}
\item with PCR5: $\Delta(\theta_6\cup\theta_7\cup \theta_8)=0.133366$
$$P(\theta_6\cup\theta_7\cup \theta_8) \in [0.866634,1]$$
$$P(\overline{\theta_6\cup\theta_7\cup \theta_8}) \in [ 0,0.133366]$$
\item with PCR6:  $\Delta(\theta_6\cup\theta_7\cup \theta_8)=0.130503$
$$P(\theta_6\cup\theta_7\cup \theta_8)  \in [ 0.869497,1 ]$$
$$P(\overline{\theta_6\cup\theta_7\cup \theta_8})  \in [0,0.130503]$$
\end{itemize}

\noindent
and when considering $\theta_7\cup \theta_8$ 
\begin{itemize}
\item with PCR5: $\Delta(\theta_7\cup\theta_8)=0.1537$
$$P(\theta_7\cup\theta_8) \in [ 0.2730, 0.4267]$$
$$P(\overline{\theta_7\cup\theta_8}) \in [ 0.5733,0.7270]$$
\item with PCR6:  $\Delta(\theta_7\cup\theta_8)=0.1537$
$$P(\theta_7\cup\theta_8) \in [ 0.2730, 0.4267]$$
$$P(\overline{\theta_7\cup\theta_8}) \in [ 0.5733,0.7270]$$
\end{itemize}
\noindent
and when considering $\theta_8$ only
\begin{itemize}
\item with PCR5: $\Delta(\theta_8)=0.071335$
$$P(\theta_8) \in [0.2730, 0.344335]$$
$$P(\bar{\theta}_8) \in [ 0.655665,0.7270]$$
\item with PCR6:  $\Delta(\theta_8)=0.076345$
$$P(\theta_8) \in [ 0.2730, 0.349345 ]$$
$$P(\bar{\theta}_8) \in [0.650655,0.7270]$$
\end{itemize}

Based on DSmP transformation, one gets a pretty low probability on $\theta_8$ and on $\theta_7\cup\theta_8$, but a very high probability on the most prudent decision-support fypothesis $\theta_6\cup\theta_7\cup\theta_8$ because
\begin{align*}
DSmP_{\epsilon,PCR5}(\theta_6\cup\theta_7\cup\theta_8)&= 0.9992\\
DSmP_{\epsilon,PCR6}(\theta_6\cup\theta_7\cup\theta_8)&= 0.9993\\   
DSmP_{\epsilon,PCR5}(\theta_7\cup\theta_8)&= 0.3168\\
DSmP_{\epsilon,PCR6}(\theta_7\cup\theta_8)&= 0.3180\\
DSmP_{\epsilon,PCR5}(\theta_8)&= 0.3166 \\
DSmP_{\epsilon,PCR6}(\theta_8)&=  0.3178
\end{align*}

\begin{itemize}
\item[-] {\bf{Answer to Q1:}} Based on these results, one sees that the decision based either on the max of credibility, the max of plausibility or the max of DSmP considering both cases $\theta_8$ or $\theta_7\cup\theta_8$ is  to: NOT Evacuate the building $B$, whereas the most prudent/cautious strategy suggests the opposite, i.e. the evacuation of the building $B$.
\end{itemize}

Let's examine the results of fusion systems $m_0\oplus m_2$ and $m_0\oplus m_1\oplus m_3$ corresponding to the input Table \ref{Table4}.
Naturally, one gets same results for the fusion $m_0\oplus m_2$ as in Example 3 and for the fusion $m_0\oplus m_1\oplus m_3$ one gets:

 \begin{table}[!h]
 \small
\centering
 \begin{tabular}{|l|c|c|}
    \hline
    focal element     & $m_{PCR5}(.)$  & $m_{PCR6}(.)$ \\
    \hline  
       $\theta_8$ & 0.091 &  0.091\\
       $\theta_4\cup\theta_8$ & 0.009 & 0.009 \\    
       $\theta_6\cup\theta_8$ & 0.819 & 0.819 \\    
        $I_t$ & 0.081 & 0.081 \\    
     \hline
  \end{tabular}
  \caption{Result of $m_0\oplus m_1\oplus m_3$.}
\label{ResultExample4m0m1m3}
\end{table}

\begin{table}[!h]
 \small
\centering
 \begin{tabular}{|l|c|c|}
    \hline
 Singletons       & $DSmP_{\epsilon,PCR5}(.)$  & $DSmP_{\epsilon,PCR6}(.)$ \\
    \hline 
$\theta_1$ &      0.0008  &  0.0008\\
$\theta_2$ &      0.0008  &  0.0008\\
$\theta_3$ &      0.0008   & 0.0008\\
$\theta_4$ &      0.0009   & 0.0009\\
$\theta_5$ &      0.0008   & 0.0008\\
$\theta_6$ &      0.0096  &  0.0096\\
$\theta_7$ &     0.0008   & 0.0008\\
$\theta_8$ &      0.9854  &  0.9854\\
       \hline
  \end{tabular}
  \caption{$DSmP_{\epsilon}$ of $m_0\oplus m_1\oplus m_3$.}
\label{ResultExample4m0m1m3DSmP}
\end{table}

\begin{table}[!h]
 \small
\centering
 \begin{tabular}{|l|c|c|}
    \hline
 Singletons       & $BetP_{PCR5}(.)$  & $BetP_{PCR6}(.)$ \\
    \hline 
$\theta_1$ &      0.0101 &   0.0101\\
$\theta_2$ &      0.0101  &  0.0101\\
$\theta_3$ &      0.0101 &   0.0101\\
$\theta_4$ &     0.0146  &  0.0146\\
$\theta_5$ &      0.0101  &  0.0101\\
$\theta_6$ &      0.4196   &  0.4196\\
$\theta_7$ &     0.0101 &   0.0101\\
$\theta_8$ &      0.5151 &   0.5151\\
       \hline
  \end{tabular}
  \caption{$BetP$ of $m_0\oplus m_1\oplus m_3$.}
\label{ResultExample4m0m1m3BetP}
\end{table}    

As in Example 3, based on $m_0\oplus m_2$ fusion result, one gets a large imprecision on evaluation of probabilities of decision-support hypotheses since for $\theta_6\cup\theta_7\cup\theta_8$, one has
$\Delta_{02}(\theta_6\cup\theta_7\cup\theta_8)=1$ and
$$P(\theta_6\cup\theta_7\cup\theta_8)\in [0,1]$$
$$P(\overline{\theta_6\cup\theta_7\cup\theta_8})\in [0, 1]$$
\noindent
for $\theta_7\cup\theta_8$, one has also $\Delta_{02}(\theta_7\cup\theta_8)=1$ with
$$P(\theta_7\cup\theta_8)\in [0,1]$$
$$P(\overline{\theta_7\cup\theta_8})\in [0, 1]$$
\noindent
and for $\theta_8$, one gets $\Delta_{02}(\theta_8)=  0.30875$ with
$$P(\theta_8)\in [0, 0.30875]$$
$$P(\bar{\theta}_8)\in [0.69125, 1]$$

One sees that it is impossible to take a decision when considering only $\theta_6\cup\theta_7\cup\theta_8$ or $\theta_7\cup\theta_8$ because of full imprecision of the corresponding probabilities.
Based on max of Bel or max of Pl criteria on $\theta_8$ the decision using $m_0\oplus m_2$ is to NOT evacuate the building $B$. According to Tables \ref{ResultExample3m0m2DSmP}-\ref{ResultExample3m0m2BetP}, an ambiguity appears in decision-making between $\theta_8$ and $\theta_4$ since they have the same DSmP (or BetP) values. \\

Based on $m_0\oplus m_1\oplus m_3$ fusion result, one gets $\Delta_{013}(\theta_6\cup\theta_7\cup\theta_8)=0.09$ for $\theta_6\cup\theta_7\cup\theta_8$ with
$$P(\theta_6\cup\theta_7\cup\theta_8)\in [0.91, 1]$$
$$P(\overline{\theta_6\cup\theta_7\cup\theta_8})\in [0,0.09]$$
\noindent
and $\Delta_{013}(\theta_7\cup\theta_8)=0.9090$ for $\theta_7\cup\theta_8$ with
$$P(\theta_7\cup\theta_8)\in [0.091, 1]$$
$$P(\overline{\theta_7\cup\theta_8})\in [0,0.9090]$$
\noindent
and $\Delta_{013}(\theta_8)=0.9090$ for only $\theta_8$ with
$$P(\theta_8)\in [0.091, 1]$$
$$P(\bar{\theta}_8)\in [0,0.9090]$$

\noindent
Based on max of Bel or max of Pl criteria on either $\theta_8$, $\theta_7\cup\theta_8$ or $\theta_6\cup\theta_7\cup\theta_8$ the decision using $m_0\oplus m_1\oplus m_3$ must be the evacuation of the building $B$. Same decision is drawn using DSmP or BetP results according to Tables \ref{ResultExample4m0m1m3DSmP} and  \ref{ResultExample4m0m1m3BetP}. 

\begin{itemize}
\item[-]{\bf{Answer to Q2:}} Similar remarks and conclusions to those given in Example 3 held also for Example 4, i.e. it is better to adopt the most prudent strategy (i.e. to consider $\theta_6\cup\theta_7\cup\theta_8$ as decision-support hypothesis) and to trust the most precise fusion system with respect this hypothesis, which is in this example the subsystem $m_0\oplus m_1\oplus m_3$. Based only on $m_0\oplus m_1\oplus m_3$ the final decision will be to evacuate the building $B$ when one has in hands such highly uncertain prior information $m_0$.
\end{itemize}

\subsection{Impact of no prior information}

\noindent
{\bf{Example 5}}:  Let's examine the result of the fusion process if one doesn't include\footnote{Or equivalently we can take $m_0$ as the vacuous bba corresponding to $m_0(I_t)=1$ and to the fully ignorant prior source.} the prior information $m_0(.)$ and if we combine directly only the three sources $m_1\oplus m_2 \oplus m_3$ altogether with PCR5 or PCR6.
 \begin{table}[!h]
 \small
\centering
 \begin{tabular}{|l|c|c|c|}
    \hline
    focal element      & $m_{1}(.)$ & $m_{2}(.)$ & $m_{3}(.)$ \\
    \hline  
    $\theta_4\cup\theta_8$   & 0 & 0.3  & 0 \\
    $\theta_6\cup\theta_8$ & 0.75 & 0 & 0.25\\
    $\overline{\theta_4\cup\theta_8}$  & 0 & 0.7 & 0\\
    $I_t$ & 0.25 & 0 & 0.75\\
   \hline
  \end{tabular}
  \caption{Quantitative inputs of VBIED problem.}
\label{Table5}
\end{table}
 \begin{table}[!h]
 \small
\centering
 \begin{tabular}{|l|c|c|}
    \hline
    focal element     & $m_{PCR5}(.)$  & $m_{PCR6}(.)$ \\
    \hline  
      $\theta_6$ & 0.56875 &  0.56875\\
     $\theta_1\cup\theta_2\cup\theta_3\cup\theta_5\cup\theta_6\cup\theta_7$ & 0.13125 &  0.13125\\
      $\theta_8$ & 0.24375 & 0.24375 \\
      $\theta_4\cup\theta_8$ & 0.05625 & 0.05625 \\    
       \hline
  \end{tabular}
  \caption{Result of $m_1\oplus m_2 \oplus m_3$ for Table \ref{Table5}.}
\label{ResultTable5m1m2m3}
\end{table}

   \begin{table}[!h]
 \small
\centering
 \begin{tabular}{|l|c|c|}
    \hline
 Singletons       & $DSmP_{\epsilon,PCR5}(.)$  & $DSmP_{\epsilon,PCR6}(.)$ \\
    \hline 
$\theta_1$ &      0.0002   & 0.0002\\
$\theta_2$ &      0.0002   & 0.0002\\
$\theta_3$ &      0.0002   & 0.0002\\
$\theta_4$ &      0.0002   & 0.0002\\
$\theta_5$ &      0.0002   & 0.0002\\
$\theta_6$ &       0.6989  &  0.6989\\
$\theta_7$ &       0.0002   & 0.0002\\
$\theta_8$ &       0.2998  & 0.2998\\
       \hline
  \end{tabular}
\caption{$DSmP_{\epsilon}$ of $m_1\oplus m_2\oplus m_3$ for Table \ref{Table5}.}
\label{ResultTable5m1m2m3DSmP}
\end{table}

\begin{table}[!h]
 \small
\centering
 \begin{tabular}{|l|c|c|}
    \hline
 Singletons       & $BetP_{PCR5}(.)$  & $BetP_{PCR6}(.)$ \\
    \hline 
$\theta_1$ &      0.0219  &  0.0219\\
$\theta_2$ &      0.0219  &  0.0219\\
$\theta_3$ &      0.0219  &  0.0219\\
$\theta_4$ &      0.0281  &  0.0281\\
$\theta_5$ &      0.0219   & 0.0219\\
$\theta_6$ &      0.5906  &  0.5906\\
$\theta_7$ &       0.0219 &   0.0219\\
$\theta_8$ &       0.2719 &   0.2719\\
       \hline
  \end{tabular}
\caption{$BetP$ of $m_1\oplus m_2\oplus m_3$ for Table \ref{Table5}.}
\label{ResultTable5m1m2m3BetP}
\end{table}         

\noindent
One gets $\Delta(\theta_6\cup\theta_7\cup\theta_8)=0.1875$ for $\theta_6\cup\theta_7\cup\theta_8$ with
$$P(\theta_6\cup\theta_7\cup\theta_8)\in [0.8125,1]$$
$$P(\overline{\theta_6\cup\theta_7\cup\theta_8})\in [0, 0.1875]$$
\noindent
for $\theta_7\cup\theta_8$, one has also $\Delta(\theta_7\cup\theta_8)=0.1875$ with
$$P(\theta_7\cup\theta_8)\in [0.24375,0.43125]$$
$$P(\overline{\theta_7\cup\theta_8})\in [0.56875,0.75625]$$
\noindent
and for $\theta_8$, one gets $\Delta(\theta_8)=  0.05625$ with
$$P(\theta_8)\in [0.24375,0.3], \quad P(\bar{\theta}_8)\in [0.7,0.75625]$$

The result presented in Table \ref{ResultTable5m1m2m3} is obviously the same as the one we obtain by combining the sources $m_0\oplus m_1\oplus m_2 \oplus m_3$ altogether when taking $m_0(.)$ as the vacuous belief assignment, i.e. when $m_0(I_t)=1$. 

 \begin{itemize}
\item[-] {\bf{Answer to Q1:}}  Based on results of Tables \ref{ResultTable5m1m2m3}-\ref{ResultTable5m1m2m3BetP} the decision based on max of belief, max of plausibility on either $\theta_8$ or $\theta_7\cup \theta_8$ is to NOT evacuate building $B$. Same conclusions is obtained when analyzing values of DSmP or BetP of $\theta_8$ or $\theta_7\cup \theta_8$. However, if we adopt the most prudent strategy based on decision-support hypothesis $\theta_6\cup\theta_7\cup\theta_8$ the decision will be to evacuate the building $B$ since $DSmP(\theta_6\cup\theta_7\cup\theta_8)=0.9989$. So we see the strong impact of the miss of prior information in the decision-making support process (by comparison between Example 1 and this example) when adopting more risky strategies for decision-making based either on $\theta_7\cup \theta_8$ or on $\theta_8$ only.
\end{itemize}

Let's compare now the source $m_2$ with respect to the $m_1\oplus m_3$ fusion system when no prior information $m_0$ is used. Naturally, there is no need to fusion $m_2$ since we consider it alone. One has  $\Delta_{2}(\theta_6\cup\theta_7\cup\theta_8)=\Delta_{2}(\theta_7\cup\theta_8)=1$ (i.e.  the full imprecision on $P(\theta_6\cup\theta_7\cup\theta_8)$ and on $P(\theta_7\cup\theta_8)$) whereas $\Delta_{2}(\theta_8)=  0.3$ with
$$P(\theta_8)\in [0, 0.3], \quad P(\bar{\theta}_8)\in [0.7, 1]$$
\noindent
DSmP and BetP of $m_2(.)$ are the same since there is no singleton as focal element for $m_2(.)$ - see Table \ref{ResultExample5m2DSmP}.
\begin{table}[!h]
 \small
\centering
 \begin{tabular}{|l|c|c|}
    \hline
 Singletons       & $DSmP_{\epsilon,PCR5}(.)$ & $BetP(.)$\\
    \hline 
$\theta_1$ &      0.1167 & 0.1167\\
$\theta_2$ &      0.1167 & 0.1167\\
$\theta_3$ &      0.1167 & 0.1167\\
$\theta_4$ &      0.1500 & 0.1500\\
$\theta_5$ &      0.1167 & 0.1167\\
$\theta_6$ &      0.1167 & 0.1167 \\
$\theta_7$ &      0.1167 & 0.1167 \\
$\theta_8$ &      0.1500 & 0.1500\\ 
       \hline
  \end{tabular}
  \caption{$DSmP_{\epsilon}$ and $BetP$ of $m_2$.}
\label{ResultExample5m2DSmP}
\end{table}  
    
Based on max of Bel or max of Pl criteria on $\theta_8$ (optimistic/risky strategy) the decision using $m_2$ (ANPR system alone) should be to NOT evacuate the building $B$. No decision can be taken using decision-support hypotheses $\theta_6\cup\theta_7\cup\theta_8$ or $ \theta_7\cup\theta_8$, nor on DSmP or BetP values since there is an ambiguity between $\theta_8$ and $\theta_4$. \\

Now if we combine $m_1$ with $m_3$ using PCR5 or PCR6 we get\footnote{Note that for two sources, PCR6 equals PCR5 \cite{DSmTBook1-3}, Vol.2.} results given in Tables \ref{ResultExample5m1m3} and \ref{ResultExample5m1m3DSmP}.\\

 \begin{table}[!h]
 \small
\centering
 \begin{tabular}{|l|c|c|}
    \hline
    focal element     & $m_{PCR5}(.)$  & $m_{PCR6}(.)$ \\
    \hline  
         $\theta_6\cup\theta_8$ &0.8125 & 0.8125 \\    
        $I_t$ & 0.1875 & 0.1875 \\    
     \hline
  \end{tabular}
  \caption{Result of $m_1\oplus m_3$.}
\label{ResultExample5m1m3}
\end{table}

\begin{table}[!h]
 \small
\centering
 \begin{tabular}{|l|c|c|}
    \hline
 Singletons       & $DSmP_{\epsilon,PCR5}(.)$  & $DSmP_{\epsilon,PCR6}(.)$ \\
    \hline 
$\theta_1$ &      0.0234   & 0.0234\\
$\theta_2$ &      0.0234 &   0.0234\\
$\theta_3$ &      0.0234 &   0.0234\\
$\theta_4$ &     0.0234   & 0.0234\\
$\theta_5$ &      0.0234 &   0.0234\\
$\theta_6$ &      0.4297  &  0.4297\\
$\theta_7$ &     0.0234   & 0.0234\\
$\theta_8$ &      0.4297  &  0.4297\\
       \hline
  \end{tabular}
  \caption{$DSmP_{\epsilon}$ of $m_1\oplus m_3$.}
\label{ResultExample5m1m3DSmP}
\end{table}   

The values of $BetP(.)$ are same as those of $DSmP(.)$ because there is no singleton as focal element of $m_1\oplus m_3$.\\

Whence $\Delta_{13}(\theta_6\cup\theta_7\cup\theta_8)=0.1875$ with
$$P(\theta_6\cup\theta_7\cup\theta_8)\in [0.8125,1]$$
$$P(\overline{\theta_6\cup\theta_7\cup\theta_8})\in [0,0.1875]$$
\noindent
and $\Delta_{13}(\theta_7\cup\theta_8)=1$ with
$$P(\theta_7\cup\theta_8)\in [0,1], \qquad P(\overline{\theta_7\cup\theta_8})\in [0,1]$$
\noindent
and also $\Delta_{13}(\theta_8)=1$ with
$$P(\theta_8)\in [0,1], \qquad P(\bar{\theta}_8)\in [0,1]$$

Based on max of Bel or max of Pl criteria on $\theta_6\cup\theta_7\cup\theta_8$, the decision using $m_1\oplus m_3$ must be the evacuation of the building $B$. Same decision must be drawn when using DSmP results according to Table \ref{ResultExample5m1m3DSmP}.  No decision can be drawn based only on $\theta_8$ or on $\theta_7\cup\theta_8$ because of full imprecision  on their corresponding probabilities.

\begin{itemize}
\item[-]{\bf{Answer to Q2:}} Similar remarks and conclusions to those given in Example 3 held also for Example 5, i.e. it is better to trust the most precise source for the most prudent decision-support hypothesis, and to decide to evacuate the building $B$ if one has no prior information rather than using only information based on APNR system.\\
\end{itemize}


\noindent
{\bf{Example 6}}:  It can be easily verified that the same analysis, remarks and conclusions for Q1 and Q2 as for Example 5 also hold when considering the sources $m_1$ and $m_3$ corresponding to the following input Table

 \begin{table}[!h]
 \small
\centering
 \begin{tabular}{|l|c|c|c|}
    \hline
    focal element      & $m_{1}(.)$ & $m_{2}(.)$ & $m_{3}(.)$ \\
    \hline  
    $\theta_4\cup\theta_8$   & 0 & 0.3  & 0 \\
     $\theta_6\cup\theta_8$  & 0.90 & 0 & 0.10\\
    $\overline{\theta_4\cup\theta_8}$  & 0 & 0.7 & 0\\
    $I_t$ & 0.10 & 0 & 0.90\\
   \hline
  \end{tabular}
  \caption{Quantitative inputs of VBIED problem.}
\label{Table6}
\end{table}

\subsection{Impact of reliability of sources}

The reliability of sources (when known) can be easily taken into using Shafer's classical discounting technique \cite{Shafer_1976}, p. 252, which consists in multiplying the masses of focal elements by the reliability factor $\alpha$, and transferring all the remaining discounted mass to the full ignorance $\Theta$. When $\alpha < 1$, such very simple reliability discounting technique discounts all focal elements with the same factor $\alpha$ and it increases the non specificity of the discounted sources since the mass committed to the full ignorance always increases.
When $\alpha=1$, no reliability discounting occurs (the bba is kept unchanged). Mathematically, Shafer's discounting technique for taking into account the reliability factor $\alpha \in [0,1]$ of a given source with a bba $m(.)$ and a frame $\Theta$ is defined by:
\begin{equation}
\begin{cases}
m_{\alpha}(X)=\alpha \cdot m(X), \quad \text{for}\ X\neq\Theta\\
m_{\alpha}(\Theta)=\alpha \cdot m(\Theta) + (1-\alpha)
\end{cases}
\label{eq:reliabilitydiscounting}
\end{equation}


\noindent
{\bf{Example 7}}:  Let's consider back the inputs of Table \ref{Table4}. The impact of strong unreliability of prior information $m_0$ has already been analyzed in Examples 3 and 4 by considering actually $\alpha_0=0.1$. Here we analyze the impact of reliabilities of sources $m_0$, $m_1$, $m_2$ and $m_3$ according presentation done in section 2.4 and we choose the following set of reliability factors $\alpha_0=0.9$, $\alpha_1=0.75$, $\alpha_2=0.75$ and $\alpha_3=0.25$. These values have been chosen approximatively but they reflect the fact that one has a very good confidence in our prior information, a good confidence in sources 1 and 2, and a low confidence in source 3.  Let's examine the change in the fusion result of sources. Applying reliability discounting technique \cite{Shafer_1976}, the new inputs corresponding to the discounted bba's by \eqref{eq:reliabilitydiscounting} are given in Table \ref{Table7}.\\

 \begin{table}[!h]
 \small
\centering
 \begin{tabular}{|l|c|c|c|c|}
    \hline
    focal element     & $m_{\alpha_0}(.)$  & $m_{\alpha_1}(.)$ & $m_{\alpha_2}(.)$ & $m_{\alpha_3}(.)$ \\
    \hline  
    $\theta_4\cup\theta_8$       & 0.90  & 0    & 0.2250  & 0 \\
    $\theta_6\cup\theta_8$       & 0       & 0.5625 & 0     & 0.0625\\
    $\overline{\theta_4\cup\theta_8}$       & 0       & 0    & 0.5250  & 0\\
    $I_t$ & 0.10 & 0.4375 & 0.2500    & 0.9375\\
   \hline
  \end{tabular}
  \caption{Discounted inputs with $\alpha_0=0.9$, $\alpha_1=0.75$, $\alpha_2=0.75$ and $\alpha_3=0.25$.}
\label{Table7}
\end{table}

 \begin{table}[!h]
 \small
\centering
 \begin{tabular}{|l|c|c|}
    \hline
    focal element     & $m_{PCR5}(.)$  & $m_{PCR6}(.)$ \\
    \hline  
     $\theta_6$ & 0.030967 &  0.030967\\
     $\overline{\theta_4\cup\theta_8}$ & 0.13119 &  0.11037\\
      $\theta_8$ & 0.26543 & 0.26543 \\
       $\theta_4\cup\theta_8$ & 0.37256 & 0.33686 \\    
       $\theta_6\cup\theta_8$ & 0.063483 & 0.068147\\    
           $I_t$ & 0.13637 & 0.18822 \\      
       \hline
  \end{tabular}
  \caption{Result of $m_0\oplus m_1\oplus m_2\oplus m_3$ for Table \ref{Table7}.}
\label{ResultTable7}
\end{table}
    
 \begin{table}[!h]
 \small
\centering
 \begin{tabular}{|l|c|c|}
    \hline
 Singletons       & $DSmP_{\epsilon,PCR5}(.)$  & $DSmP_{\epsilon,PCR6}(.)$ \\
    \hline 
$\theta_1$ &      0.0040  &  0.0036\\
$\theta_2$ &      0.0040  &  0.0036\\
$\theta_3$ &      0.0040  &  0.0036\\
$\theta_4$ &      0.0018  &  0.0019\\
$\theta_5$ &      0.0040  &   0.0036\\
$\theta_6$ &      0.1655  &  0.1535\\
$\theta_7$ &      0.0040  &  0.0036\\
$\theta_8$ &      0.8126  &  0.8266\\
       \hline
  \end{tabular}
 \caption{$DSmP_{\epsilon}$ of $m_0\oplus m_1\oplus m_2\oplus m_3$ for Table \ref{Table7}.}
\label{ResultTable7DSmP}
\end{table}

         \begin{table}[!h]
 \small
\centering
 \begin{tabular}{|l|c|c|}
    \hline
 Singletons       & $BetP_{PCR5}(.)$  & $BetP_{PCR6}(.)$ \\
    \hline 
$\theta_1$ &      0.0389 &   0.0419\\
$\theta_2$ &      0.0389  &  0.0419\\
$\theta_3$ &      0.0389  &  0.0419\\
$\theta_4$ &      0.2033  &  0.1920\\
$\theta_5$ &      0.0389 &   0.0419\\
$\theta_6$ &      0.1016  &  0.1070\\
$\theta_7$ &       0.0389 &  0.0419\\
$\theta_8$ &      0.5005  &  0.4915\\
       \hline
  \end{tabular}
 \caption{$BetP$ of $m_0\oplus m_1\oplus m_2\oplus m_3$ for Table \ref{Table7}.}
\label{ResultTable7BetP}
\end{table}

From the fusion result of Table \ref{ResultTable7}, one gets when considering $\theta_6\cup\theta_7\cup \theta_8$:
\begin{itemize}
\item with PCR5: $\Delta(\theta_6\cup\theta_7\cup\theta_8)=0.64012$
$$P(\theta_6\cup\theta_7\cup\theta_8) \in [ 0.35988,1]$$
$$P(\overline{\theta_6\cup\theta_7\cup\theta_8}) \in [ 0,0.64012]$$
\item with PCR6:  $\Delta(\theta_6\cup\theta_7\cup\theta_8)=0.635456$
$$P(\theta_6\cup\theta_7\cup\theta_8) \in [ 0.364544,1] $$
$$P(\overline{\theta_6\cup\theta_7\cup\theta_8}) \in [ 0,0.635456]$$
\end{itemize}
\noindent
and when considering $\theta_7\cup \theta_8$ 
\begin{itemize}
\item with PCR5: $\Delta(\theta_7\cup\theta_8)=0.703603$
$$P(\theta_7\cup\theta_8) \in [ 0.26543,0.969033]$$
$$P(\overline{\theta_7\cup\theta_8}) \in [ 0.030967,0.73457]$$
\item with PCR6:  $\Delta(\theta_7\cup\theta_8)=0.703603$
$$P(\theta_7\cup\theta_8) \in [ 0.26543,0.969033]$$
$$P(\overline{\theta_7\cup\theta_8}) \in [ 0.030967,0.73457]$$
\end{itemize}
\noindent
and when considering $\theta_8$ only
\begin{itemize}
\item with PCR5: $\Delta(\theta_8)=0.572413$
$$P(\theta_8) \in [0.26543, 0.837843]$$
$$P(\bar{\theta}_8) \in [ 0.162157,0.73457]$$
\item with PCR6:  $\Delta(\theta_8)=0.593233$
$$P(\theta_8) \in [ 0.26543, 0.858663]$$
$$P(\bar{\theta}_8) \in [0.141337,0.73457]$$
\end{itemize}

Using DSmP transformation, one gets high probabilities in $\theta_6\cup\theta_7\cup\theta_8$, $\theta_7\cup\theta_8$ and in $\theta_8$ because
\begin{align*}
DSmP_{\epsilon,PCR5}(\theta_6\cup\theta_7\cup\theta_8)&=0.9821\\
DSmP_{\epsilon,PCR6}(\theta_6\cup\theta_7\cup\theta_8)&=0.9837\\    
DSmP_{\epsilon,PCR5}(\theta_7\cup\theta_8)&=0.8166\\
DSmP_{\epsilon,PCR6}(\theta_7\cup\theta_8)&=0.8302\\
DSmP_{\epsilon,PCR5}(\theta_8)&= 0.8126 \\
DSmP_{\epsilon,PCR6}(\theta_8)&= 0.8266
\end{align*}

\begin{itemize}
\item[-] {\bf{Answer to Q1:}} Based on these results, one sees that the decision to take is to evacuate the building $B$ since one gets a high probability in decision-support hypotheses. Same conclusion is drawn when using max of Bel of max of Pl criteria. So there is a little impact of reliability discounting on the final decision with respect to Example 1. It is however worth to note that introducing reliability discounting increases the non specificity of information since now $I_t$ is a new focal element of $m_{\alpha_0}$ and of $m_{\alpha_2}$ and in the final result we get the new focal element $\theta_6$ appearing with PCR5 or PCR6 fusion rules. This $\theta_6$ focal element doesn't exist in Example 1 when no reliability discounting is used. The decision to take in this case is  to: {\bf{Evacuate the building $B$}}.
\end{itemize}
    
To answer to the question Q2 for this Example 7, let's compute the fusion results of the fusion $m_0\oplus m_2$ and  $m_0\oplus m_1\oplus m_3$ using inputs given in Table \ref{Table7}. The fusion results with corresponding DSmP are given in the Tables \ref{ResultExample7m0m2}-\ref{ResultExample7m0m1m3DSmP}.\\

\begin{table}[!h]
 \small
\centering
 \begin{tabular}{|l|c|c|}
    \hline
    focal element     & $m_{PCR5}(.)$  & $m_{PCR6}(.)$ \\
    \hline  
       $\theta_4\cup\theta_8$ & 0.74842 & 0.74842 \\    
    $\overline{\theta_4\cup\theta_8}$ & 0.22658 &  0.22658\\
        $I_t$ & 0.025 & 0.025 \\    
      \hline
  \end{tabular}
  \caption{Result of $m_0\oplus m_2$.}
\label{ResultExample7m0m2}
\end{table}

\begin{table}[!h]
 \small
\centering
 \begin{tabular}{|l|c|c|}
    \hline
 Singletons       & $DSmP_{\epsilon,PCR5}(.)$  & $DSmP_{\epsilon,PCR6}(.)$ \\
    \hline 
$\theta_1$ &      0.0409  &  0.0409\\
$\theta_2$ &      0.0409  &  0.0409\\
$\theta_3$ &      0.0409  &  0.0409\\
$\theta_4$ &      0.3773  & 0.3773\\
$\theta_5$ &      0.0409  &  0.0409\\
$\theta_6$ &      0.0409  &  0.0409\\
$\theta_7$ &      0.0409  &  0.0409\\
$\theta_8$ &      0.3773  &  0.3773\\
       \hline
  \end{tabular}
  \caption{$DSmP_{\epsilon}$ of $m_0\oplus m_2$.}
\label{ResultExample7m0m2DSmP}
\end{table}    

The result of BetP transformation is the same as with DSmP transformation since there is no singleton element as focal element of the resulting bba's when using PCR5 or PCR6 fusion rule.\\

Based on $m_0\oplus m_2$ fusion result, one gets a large imprecision\footnote{This imprecision is larger than in Example 1 which is normal because one has degraded the information of both prior and the source $m_2$.} on $P(\theta_8)$ since $\Delta_{02}(\theta_8)= 0.77342$ with
$$P(\theta_8)\in [0,0.77342]$$
$$P(\bar{\theta}_8)\in [0.22658, 1]$$
\noindent
and one gets a total imprecision when considering either $\theta_7\cup\theta_8$ or  $\theta_6\cup\theta_7\cup\theta_8$ since
$$P(\theta_7\cup\theta_8)\in [0,1], \quad P(\overline{\theta_7\cup\theta_8})\in [0, 1]$$
$$P(\theta_6\cup\theta_7\cup\theta_8)\in [0,1], \quad P(\overline{\theta_6\cup\theta_7\cup\theta_8})\in [0, 1]$$

Based on max of bel or max of Pl criteria the decision using $m_0\oplus m_2$ (i.e. with discounted sources $m_0$ and ANPR system $m_2$) should be to NOT evacuate the building $B$ when working with decision-support hypothesis $\theta_8$. No clear decision can be taken when working with $\theta_6\cup\theta_7\cup\theta_8$ or $\theta_7\cup\theta_8$. Ambiguity in decision-making occurs between $\theta_8=(A,V,B)$ and $\theta_4=(A,V,\bar{B})$ when using DSmP or BetP transformations.\\

Let's examine now the result of the   $m_0\oplus m_1\oplus m_3$ fusion given in Tables \ref{ResultExample7m0m1m3} and \ref{ResultExample7m0m1m3DSmP}.
 \begin{table}[!h]
 \small
\centering
 \begin{tabular}{|l|c|c|}
    \hline
    focal element     & $m_{PCR5}(.)$  & $m_{PCR6}(.)$ \\
    \hline  
       $\theta_8$ & 0.53086 &  0.53086\\
       $\theta_4\cup\theta_8$ & 0.36914 & 0.36914\\   
       $\theta_6\cup\theta_8$ & 0.058984 & 0.058984\\   
        $I_t$ & 0.041016 & 0.041016\\   
       \hline
  \end{tabular}
  \caption{Result of $m_0\oplus m_1\oplus m_3$.}
\label{ResultExample7m0m1m3}
\end{table}
   
\begin{table}[!h]
 \small
\centering
 \begin{tabular}{|l|c|c|}
    \hline
 Singletons       & $DSmP_{\epsilon,PCR5}(.)$  & $DSmP_{\epsilon,PCR6}(.)$ \\
    \hline 
$\theta_1$ &     0.0001   & 0.0001\\
$\theta_2$ &      0.0001  & 0.0001\\
$\theta_3$ &      0.0001  & 0.0001\\
$\theta_4$ &      0.0008  &  0.0008\\
$\theta_5$ &      0.0001  &  0.0001\\
$\theta_6$ &      0.0002  &  0.0002\\
$\theta_7$ &      0.0001  &  0.0001\\
$\theta_8$ &      0.9987  &  0.9987\\
       \hline
  \end{tabular}
  \caption{$DSmP_{\epsilon}$ of $m_0\oplus m_1\oplus m_3$.}
\label{ResultExample7m0m1m3DSmP}
\end{table}

\begin{table}[!h]
 \small
\centering
 \begin{tabular}{|l|c|c|}
    \hline
 Singletons       & $BetP_{PCR5}(.)$  & $BetP_{PCR6}(.)$ \\
    \hline 
$\theta_1$ &     0.0051   & 0.0051\\
$\theta_2$ &     0.0051   & 0.0051\\
$\theta_3$ &      0.0051   & 0.0051\\
$\theta_4$ &      0.1897 &  0.1897\\
$\theta_5$ &      0.0051   & 0.0051\\
$\theta_6$ &      0.0346 &  0.0346\\
$\theta_7$ &      0.0051   & 0.0051\\
$\theta_8$ &      0.7500  &  0.7500\\
       \hline
  \end{tabular}
  \caption{$BetP$ of $m_0\oplus m_1\oplus m_3$.}
\label{ResultExample7m0m1m3BetP}
\end{table}

One sees clearly the impact of reliability discounting on the specificity of information provided by the fusion of sources.
Indeed when using the discounting of sources (mainly because we introduce $I_t$ as focal element for $m_0$) one gets now 4 focal elements whereas we did get only two focal elements when no discounting was used (see Table \ref{ResultExample1m0m1m3DSmP}). Based on $m_0\oplus m_1\oplus m_3$ fusion result, one gets $\Delta_{013}(\theta_6\cup\theta_7\cup \theta_8)= 0.410156$ with
$$P(\theta_6\cup\theta_7\cup\theta_8)\in [0.589844, 1]$$
$$P(\overline{\theta_6\cup\theta_7\cup\theta_8})\in [0,0.410156]$$
\noindent
and also $\Delta_{013}(\theta_7\cup \theta_8)= 0.46914$ with
$$P(\theta_7\cup\theta_8)\in [0.53086, 1]$$
$$P(\overline{\theta_7\cup\theta_8})\in [0,0.46914]$$
\noindent
and $\Delta_{013}(\theta_8)= 0.46914$ with
$$P(\theta_8)\in [0.53086, 1]$$
$$P(\bar{\theta}_8)\in [0,0.46914]$$

Based on max of Bel or max of Pl, the decision using $m_0\oplus m_1\oplus m_3$ should be to evacuate the building $B$. Same decision would be taken based on DSmP values. It is worth to note that the precision on the result obtained with  $m_0\oplus m_1\oplus m_3$ is much better than with $m_0\oplus m_2$ since $\Delta_{013}(\theta_8) < \Delta_{02}(\theta_8)$, or $\Delta_{013}(\theta_7\cup\theta_8) < \Delta_{02}(\theta_7\cup\theta_8)$,  or $\Delta_{013}(\theta_6\cup\theta_7\cup\theta_8) < \Delta_{02}(\theta_6\cup\theta_7\cup\theta_8)$. Moreover it is easy to verify that $m_0\oplus m_1\oplus m_3$ fusion system is more informative than $m_0\oplus m_2$ fusion system because Shannon entropy of DSmP of $m_0\oplus m_2$ is much bigger than Shannon entropy of DSmP of $m_0\oplus m_1\oplus m_3$.

\begin{itemize}
\item[-]{\bf{Answer to Q2:}}  Since the information obtained by the fusion $m_0\oplus m_2$ is less informative and less precise than the information obtained with the fusion $m_0\oplus m_1\oplus m_3$, it is better to choose and to trust the fusion system $m_0\oplus m_1\oplus m_3$ rather than $m_0\oplus m_2$. Based on this choice, the final decision will be to evacuate the building $B$.\\
\end{itemize}
    
\subsection{Impact of importance of sources}

The importance discounting technique has been proposed recently by the authors in \cite{FSJDJMT2010} and consists in discounting the masses of focal elements by a factor $\beta\in[0,1]$ and in transferring the remaining mass to empty set, i.e.
\begin{equation}
\begin{cases}
m_{\beta }(X)=\beta \cdot m(X), \quad \text{for}\ X\neq\emptyset\\
m_{\beta}(\emptyset)=\beta \cdot  m(\emptyset) + (1-\beta)
\end{cases}
\label{eq:importancediscounting}
\end{equation}
\noindent
It has been proved in \cite{FSJDJMT2010} that such importance discounting technique preserves the specificity of the information and that Dempster's rule of combination doesn't respond to such new interesting discounting technique specially useful and crucial in multicriteria decision-making support. \\

In the extreme case, the method proposed in \cite{FSJDJMT2010} reinforces the highest mass of the focal element of the source having the biggest importance factor as soon as the other sources have their importance factors tending towards zero. This reinforcement may be a disputable behavior. To avoid such behavior, we propose here to use the same importance discounting technique, but the fusion of discounted sources is done a bit differently in three steps as follows:
\begin{itemize}
\item Step 1: Discount each source with its importance discounting factor according to \eqref{eq:importancediscounting}.
\item Step 2: Apply PCR5 or PCR6 fusion rule with unnormalized discounted bba's, i.e. as if the discounted mass committed to empty set for each source was zero.
\item Step 3: Normalize the result to get the sum of masses of focal elements to be one.
\end{itemize}

Let's examine the impact of the importance of the sources in the fusion process for final decision-making through the next very simple illustrating example.\\


\noindent
{\bf{Example 8}}:  To evaluate this we consider the same inputs as in Table \ref{Table1} and we consider that source 1 (Analyst 1 with 10 years experience) is much more important than source 3 (Analyst 2 with no experience). To reflect the difference between importance of this sources we consider the following relative importance factors $\beta_1=0.9$ and $\beta_3=0.5$. We also assume that source 0 (prior information) and source 2 (ANPR system) have the same maximal importance, i.e. $\beta_0=\beta_2=1$, i.e $\beta=(\beta_0,\beta_1,\beta_2,\beta_3)=(1,0.9,1,0.5)$. These values have been chosen approximatively  but they do reflect the fact that sources $m_0$ and $m_2$ have same importance in the fusion process, and that sources $m_1$ and $m_3$ may have less importance in the fusion process taking into the fact that $m_3$ is considered as less important than $m_1$. Let's examine the change in the fusion result of sources in this example with respect to what we get in Example 1. \\

In applying importance discounting technique \cite{FSJDJMT2010} with the aforementioned fusion approach, the new inputs corresponding to the discounted (unnormalized) bba's by \eqref{eq:importancediscounting} are given in Table \ref{Table8bis}.

 \begin{table}[!h]
 \small
\centering
 \begin{tabular}{|l|c|c|c|c|}
    \hline
    focal element     & $m_{\beta_0}(.)$  & $m_{\beta_1}(.)$ & $m_{\beta_2}(.)$ & $m_{\beta_3}(.)$ \\
    \hline  
    $\emptyset$       & 0  & 0.1    & 0       & 0.5 \\
    $\theta_4\cup\theta_8$        & 1  & 0       & 0.70  & 0 \\
    $\theta_6\cup\theta_8$       & 0  & 0.675    & 0       & 0.125\\
    $\overline{\theta_4\cup\theta_8}$        & 0  & 0       & 0.30  & 0\\
    $I_t$ & 0 & 0.225  & 0       & 0.375\\
   \hline
  \end{tabular}
  \caption{Discounted inputs with $\beta_0=1$, $\beta_1=0.9$, $\beta_2=1$ and $\beta_3=0.5$.}
\label{Table8bis}
\end{table}
\noindent

\noindent
and the fusion result is given in Table \ref{ResultTable8bisnormalizedV2}.

 \begin{table}[!h]
 \small
\centering
 \begin{tabular}{|l|c|c|}
    \hline
    focal element  & $m_{PCR5}(.)$  & $m_{PCR6}(.)$ \\
    \hline  
      $\theta_8$ & 0.24375 &  0.24375\\
       $\theta_4\cup\theta_8$ & 0.36788 &  0.33034\\    
       $\theta_6\cup\theta_8$ & 0.10552 & 0.14132\\    
     $\overline{\theta_4\cup\theta_8}$ &0.21814 &  0.19186\\
           $I_t$ & 0.064701 & 0.092734 \\      
       \hline
  \end{tabular}
  \caption{Result of $m_{\beta_0}\oplus m_{\beta_1}\oplus m_{\beta_2}\oplus m_{\beta_3}$.}
\label{ResultTable8bisnormalizedV2}
\end{table}

As we can see, the importance discounting doesn't degrade the specificity of sources since no mass is committed to partial ignorances, and it doesn't also increase the number of focal elements of the resulting bba contrariwise to the reliability discounting approach. Indeed in Table \ref{ResultTable8bisnormalizedV2} one gets only 5 focal elements whereas one gets 6 focal elements with reliability discounting as shown in Table \ref{ResultTable7} of Example 7. The corresponding DSmP and BetP values of bba's given in Table \ref{ResultTable8bisnormalizedV2} are summarized in Tables \ref{ResultTable8bisDSmP} and \ref{ResultTable8bisBetP} .\\     
    
 \begin{table}[!h]
 \small
\centering
 \begin{tabular}{|l|c|c|}
    \hline
 Singletons       & $DSmP_{\epsilon,PCR5}(.)$  & $DSmP_{\epsilon,PCR6}(.)$ \\
    \hline 
$\theta_1$ &      0.0366   & 0.0323\\
$\theta_2$ &     0.0366   & 0.0323\\
$\theta_3$ &     0.0366  &  0.0323\\
$\theta_4$ &      0.0018  &  0.0017\\
$\theta_5$ &     0.0366 &   0.0323\\
$\theta_6$ &    0.0370   & 0.0329\\
$\theta_7$ &       0.0366 &    0.0323\\
$\theta_8$ &       0.7781 &   0.8036\\
       \hline
  \end{tabular}
 \caption{$DSmP_{\epsilon}$ of $m_{\beta_0}\oplus m_{\beta_1}\oplus m_{\beta_2}\oplus m_{\beta_3}$.}
\label{ResultTable8bisDSmP}
\end{table}       

\begin{table}[!h]
 \small
\centering
 \begin{tabular}{|l|c|c|}
    \hline
 Singletons       & $BetP_{PCR5}(.)$  & $BetP_{PCR6}(.)$ \\
    \hline 
$\theta_1$ &     0.0444   & 0.0436\\
$\theta_2$ &     0.0444&    0.0436\\
$\theta_3$ &     0.0444 &   0.0436\\
$\theta_4$ &       0.1920 &   0.1768\\
$\theta_5$ &     0.0444  &  0.0436\\
$\theta_6$ &    0.0972 &   0.1142\\
$\theta_7$ &       0.0444 &   0.0436\\
$\theta_8$ &       0.4885 &   0.4912\\
       \hline
  \end{tabular}
 \caption{$BetP$ of $m_{\beta_0}\oplus m_{\beta_1}\oplus m_{\beta_2}\oplus m_{\beta_3}$.}
\label{ResultTable8bisBetP}
\end{table}       

\noindent
From the fusion result of Table \ref{ResultTable8bisnormalizedV2}, one gets when considering $\theta_6\cup\theta_7\cup \theta_8$ 
\begin{itemize}
\item with PCR5: $\Delta(\theta_6\cup\theta_7\cup\theta_8)=0.65073$
$$P(\theta_6\cup\theta_7\cup\theta_8) \in [ 0.34927,1]$$
$$P(\overline{\theta_6\cup\theta_7\cup\theta_8}) \in [ 0,0.65073]$$
\item with PCR6:  $\Delta(\theta_6\cup\theta_7\cup\theta_8)=0.61493$
$$P(\theta_6\cup\theta_7\cup\theta_8) \in [ 0.38507,1] $$
$$P(\overline{\theta_6\cup\theta_7\cup\theta_8}) \in [ 0,0.61493]$$
\end{itemize}
\noindent
and when considering $\theta_7\cup \theta_8$ 
\begin{itemize}
\item with PCR5: $\Delta(\theta_7\cup\theta_8)=0.75625$
$$P(\theta_7\cup\theta_8) \in [ 0.24375,1]$$
$$P(\overline{\theta_7\cup\theta_8}) \in [ 0,0.75625]$$
\item with PCR6:  $\Delta(\theta_7\cup\theta_8)=0.75625$
$$P(\theta_7\cup\theta_8) \in [ 0.24375,1] $$
$$P(\overline{\theta_7\cup\theta_8}) \in [ 0,0.75625]$$
\end{itemize}
\noindent
and when considering $\theta_8$ only
\begin{itemize}
\item with PCR5: $\Delta(\theta_8)=0.53811$
$$P(\theta_8) \in [0.24375,0.78186]$$
$$P(\bar{\theta}_8) \in [ 0.21814,0.75625]$$
\item with PCR6:  $\Delta(\theta_8)=0.564393$
$$P(\theta_8) \in [0.24375, 0.80814]$$
$$P(\bar{\theta}_8) \in [0.19186,0.75625]$$
\end{itemize}

Using DSmP transformation, one gets a high probability in decision-support hypotheses because
\begin{align*}
DSmP_{\epsilon,PCR5}(\theta_6\cup\theta_7\cup\theta_8)&=0.8517\\
DSmP_{\epsilon,PCR6}(\theta_6\cup\theta_7\cup\theta_8)&=0.8688\\    
DSmP_{\epsilon,PCR5}(\theta_7\cup\theta_8)&=0.8147\\
DSmP_{\epsilon,PCR6}(\theta_7\cup\theta_8)&=0.8359\\ 
DSmP_{\epsilon,PCR5}(\theta_8)&=0.7781 \\
DSmP_{\epsilon,PCR6}(\theta_8)&= 0.8036
\end{align*}

\begin{itemize}
\item[-] {\bf{Answer to Q1:}} Based on these results, one sees that the decision to take based either on max of Bel or max of Pl on  $\theta_6\cup\theta_7\cup\theta_8$, on $\theta_7\cup\theta_8$ or on $\theta_8$ only, or also based on DSmP, is to evacuate the building $B$.
\end{itemize}

To answer to the question Q2 for this Example 8, let's compute the fusion results of the fusion $m_{\beta_0}\oplus m_{\beta_2}$ and  $m_{\beta_0}\oplus m_{\beta_1}\oplus m_{\beta_3}$ using inputs given in Table \ref{Table8bis}. 
Because one has considered $\beta_0=\beta_2=1$, one does not actually discount sources $m_0$ and $m_2$ and therefore the $m_0\oplus m_2$ fusion results are already given in Tables \ref{ResultExample1m0m2} and \ref{ResultExample1m0m2DSmP} of Example 1. Therefore based on max of Bel or max of Pl criteria on $\theta_8$ the decision using $m_0\oplus m_2$ is to NOT evacuate the building $B$ since $P(\theta_8)\in [0,0.71176]$ and $P(\bar{\theta}_8)\in [0.28824,1]$ and $\Delta_{02}(\theta_8)=0.71176$. Same decision would be taken based on DSmP values with the $m_0\oplus m_2$ fusion sub-system. 

Let's now compute the fusion  $m_{\beta_0}\oplus m_{\beta_1}\oplus m_{\beta_3}$ with the importance discounted sources  $m_{\beta_0=1}=m_0$, $m_{\beta_1}$ and $m_{\beta_3}$. The fusion results are given in Tables \ref{ResultExample8m0m1m3}, \ref{ResultExample8m0m1m3DSmP} and \ref{ResultExample8m0m1m3BetP}.

    \begin{table}[!h]
 \small
\centering
 \begin{tabular}{|l|c|c|}
    \hline
    focal element     & $m_{PCR5}(.)$  & $m_{PCR6}(.)$ \\
    \hline  
       $\theta_8$ & 0.8125 &  0.8125\\
       $\theta_4\cup\theta_8$ & 0.1875 & 0.1875\\   
        \hline
  \end{tabular}
  \caption{Result of $m_{\beta_0}\oplus m_{\beta_1}\oplus m_{\beta_3}$.}
\label{ResultExample8m0m1m3}
\end{table}

\begin{table}[!h]
 \small
\centering
 \begin{tabular}{|l|c|c|}
    \hline
 Singletons       & $DSmP_{\epsilon,PCR5}(.)$  & $DSmP_{\epsilon,PCR6}(.)$ \\
    \hline 
$\theta_1$ &     0   & 0\\
$\theta_2$ &       0   & 0\\
$\theta_3$ &       0   & 0\\
$\theta_4$ &      0.0002 &   0.0002\\
$\theta_5$ &       0   & 0\\
$\theta_6$ &       0   & 0\\
$\theta_7$ &       0   & 0\\
$\theta_8$ &      0.9998 &   0.9998\\
       \hline
  \end{tabular}
  \caption{$DSmP_{\epsilon}$ of $m_{\beta_0}\oplus m_{\beta_1}\oplus m_{\beta_3}$.}
\label{ResultExample8m0m1m3DSmP}
\end{table}    

\begin{table}[!h]
 \small
\centering
 \begin{tabular}{|l|c|c|}
    \hline
 Singletons       & $BetP_{PCR5}(.)$  & $BetP_{PCR6}(.)$ \\
    \hline 
$\theta_1$ &     0   & 0\\
$\theta_2$ &       0   & 0\\
$\theta_3$ &       0   & 0\\
$\theta_4$ &      0.0002 &   0.0002\\
$\theta_5$ &       0   & 0\\
$\theta_6$ &       0   & 0\\
$\theta_7$ &       0   & 0\\
$\theta_8$ &      0.9998 &   0.9998\\
       \hline
  \end{tabular}
  \caption{$BetP$ of $m_{\beta_0}\oplus m_{\beta_1}\oplus m_{\beta_3}$.}
\label{ResultExample8m0m1m3BetP}
\end{table}    

Based on $m_{\beta_0}\oplus m_{\beta_1}\oplus m_{\beta_3}$ fusion result, one gets same result with PCR5 and PCR6 in this case, and
one gets $\Delta_{013}(\theta_6\cup\theta_7\cup\theta_8)=0.1875 $ with
$$P(\theta_6\cup\theta_7\cup\theta_8)\in [0.8125, 1]$$
$$P(\overline{\theta_6\cup\theta_7\cup\theta_8})\in [0,0.1875]$$
\noindent
and $\Delta_{013}(\theta_7\cup\theta_8)=0.1875$ with
$$P(\theta_7\cup\theta_8)\in [0.8125, 1]$$
$$P(\overline{\theta_7\cup\theta_8})\in [0,0.1875]$$
\noindent
and also $\Delta_{013}(\theta_8)=0.1875$ with
$$P(\theta_8)\in [0.8125, 1]$$
$$P(\bar{\theta}_8)\in [0,0.1875]$$

Based on max of Bel or max of Pl, the decision taken using $\theta_6\cup\theta_7\cup\theta_8$, $\theta_7\cup\theta_8$ or $\theta_8$ for the $m_{\beta_0}\oplus m_{\beta_1}\oplus m_{\beta_3}$ fusion sub-system should be to evacuate the building $B$. Same decision must be taken based on DSmP or BetP values. It is worth to note that the precision\footnote{see Example 1 for the numerical results of $m_{\beta_0=1}\oplus m_{\beta_2=1}$.} on the result obtained with  subsystem $m_{\beta_0}\oplus m_{\beta_1}\oplus m_{\beta_3}$ is much better than with  subsystem $m_{\beta_0=1}\oplus m_{\beta_2=1}$ since $(\Delta_{013}(\theta_8)=0.1875) < (\Delta_{02}(\theta_8)=0.71176)$, and also $(\Delta_{013}(\theta_7\cup\theta_8)=0.1875) < (\Delta_{02}(\theta_7\cup\theta_8)=1)$, and  $(\Delta_{013}(\theta_6\cup\theta_7\cup\theta_8)=0.1875) < (\Delta_{02}(\theta_6\cup\theta_7\cup\theta_8)=1)$.

\begin{itemize}
\item[-]{\bf{Answer to Q2:}}  The analysis of both fusion sub-systems $m_{\beta_0=1}\oplus m_{\beta_2=1}$ and $m_{\beta_0}\oplus m_{\beta_1}\oplus m_{\beta_3}$  shows that the $m_{\beta_0}\oplus m_{\beta_1}\oplus m_{\beta_3}$ subsystem must be chosen  because it provides the most precise results and therefor the decision will be to evacuate the building $B$ whatever the decision-support hypothesis is chosen $\theta_6\cup\theta_7\cup\theta_8$, $\theta_7\cup\theta_8$ or $\theta_8$.
\end{itemize}

\subsection{Using imprecise bba's}
     
 Let's examine the fusion result when dealing directly with imprecise bba's. We just consider here a simple imprecise example which considers both inputs of Examples 1 and 2 to generate imprecise bba's inputs.\\


\noindent
{\bf{Example 9}}: We consider the imprecise bba's according to input Table \ref{Table9}.\\

 \begin{table}[!h]
 \small
\centering
 \begin{tabular}{|l|c|c|c|c|}
    \hline
    focal element     & $m_{0}(.)$  & $m_{1}(.)$ & $m_{2}(.)$ & $m_{3}(.)$ \\
    \hline  
    $f_1=\theta_4\cup\theta_8$  & 1  & 0 & 0.3  & 0 \\
    $f_2=\theta_6\cup\theta_8$ & 0 & [0.75,0.9] & 0 & [0.10,0.25]\\
    $f_3=\overline{\theta_4\cup\theta_8}$  & 0 & 0 & 0.7 & 0\\
    $I_t$ & 0 & [0.1,0.25] & 0 & [0.75,0.9]\\
   \hline
  \end{tabular}
  \caption{Imprecise quantitative inputs for VBIED problem.}
\label{Table9}
\end{table}

\noindent
Applying the conjunctive rule, we have $1\times 2\times 2\times 2=8$ products to compute which are listed below:
  
  \begin{itemize}
  \item Product $\pi_1=1 \boxdot [0.75,0.90] \boxdot 0.3 \boxdot [0.10,0.25]$. Using operators on sets defined in \cite{DSmTBook1-3}, Vol.1, Chap. 6, one gets
   $\pi_1 = [0.75,0.90]\boxdot [0.03,0.075]=[0.0225,0.0675]$ which is committed to $f_1\cap f_2=\theta_8$.
    \item Product $\pi_2=1 \boxdot [0.75,0.90] \boxdot 0.3 \boxdot [0.75,0.90]$ is equal to
$[0.75,0.90]\boxdot [0.225,0.27]=[0.16875,0.243]$ which is also committed to $f_1\cap f_2=\theta_8$.
   \item Product $\pi_3=1\boxdot [0.75,0.90] \boxdot 0.7 \boxdot [0.10,0.25]=[0.0525,0.1575]$ corresponds to the imprecise mass of $f_1\cap f_2\cap f_3=\emptyset$ which will be redistributed back to $f_1$, $f_2$ and $f_3$ according to PCR6.
    \item Product $\pi_4=1\boxdot [0.75,0.90] \boxdot 0.7 \boxdot [0.75,0.90]=[0.39375,0.567]$ corresponds to the imprecise mass of $f_1\cap f_2\cap f_3\cap I_t=\emptyset$ which will be redistributed back to $f_1$, $f_2$, $f_3$ and $I_t$ according to PCR6.
       \item Product $\pi_5=1 \boxdot [0.10,0.25] \boxdot 0.3 \boxdot [0.10,0.25]=[0.003,0.01875]$ is committed to $f_1\cap f_2=\theta_8$.
       \item Product $\pi_6=1 \boxdot [0.10,0.25] \boxdot 0.3 \boxdot [0.75,0.90]=[0.0225,0.0675]$ is committed to $f_1$.
    \item Product $\pi_7=1\boxdot [0.10,0.25] \boxdot 0.7 \boxdot [0.10,0.25]=[0.007,0.04375]$ corresponds to the imprecise mass of $f_1\cap I_t \cap f_3\cap f_2=\emptyset$ which will be redistributed back to $f_1$, $f_2$, $f_3$ and $I_t$ according to PCR6.
    \item Product $\pi_8=1\boxdot [0.10,0.25] \boxdot 0.7 \boxdot [0.75,0.90]=[0.0525,0.1575]$ corresponds to the imprecise mass of $f_1\cap I_t \cap f_3\cap I_t=\emptyset$ which will be redistributed back to $f_1$, $f_3$ and $I_t$ according to PCR6.
  \end{itemize}
  
  We now redistribute the imprecise masses $\pi_3$, $\pi_4$, $\pi_7$ and $\pi_8$ associated with the empty set using PCR6 principle. Lets' compute the proportions of  $\pi_3$, $\pi_4$, $\pi_7$ and $\pi_8$ committed to each focal element involved in the conflict they are associated with.
  
\begin{itemize}
\item The product $\pi_3=[0.0525,0.1575]$ is distributed to $f_1$, $f_2$ and $f_3$ according to PCR6 as follows
\begin{align*}
 \frac{x_{f_1,\pi_3}}{1} & = \frac{y_{f_2,\pi_3}}{[0.75,0.90]\boxplus [0.10,0.25]} =  \frac{z_{f_3,\pi_3}}{0.7}\\
 & =\frac{\pi_3}{1\boxplus [0.75,0.90]\boxplus [0.10,0.25]\boxplus 0.7}\\
 & =\frac{[0.0525,0.1575]}{1.7\boxplus [0.85,1.15]} = \frac{[0.0525,0.1575]}{[2.55,2.85]} \\
 & = [ \frac{0.0525}{2.85},\frac{0.1575}{2.55} ]\\
 &=[0.018421,0.061765]
\end{align*}
\noindent
whence
\begin{align*}
x_{f_1,\pi_3}& =1\boxdot [0.018421,0.061765]\\
&=[0.018421,0.061765]\\
y_{f_2,\pi_3}& =[0.85,1.15]\boxdot [0.018421,0.061765]\\
& = [0.015658,0.071029 ]\\
z_{f_3,\pi_3}& =0.7\boxdot [0.018421,0.061765]\\
&=[0.012895,0.043236]
\end{align*}
\end{itemize}

\begin{itemize}
\item The product $\pi_4=[0.39375,0.567]$ is distributed to $f_1$, $f_2$, $f_3$ and $I_t$ according to PCR6 as follows
\begin{align*}
 \frac{x_{f_1,\pi_4}}{1} & = \frac{y_{f_2,\pi_4}}{[0.75,0.90]} =  \frac{z_{f_3,\pi_4}}{0.7} = \frac{w_{I_t,\pi_4}}{[0.75,0.90]}\\
 & =\frac{\pi_4}{1\boxplus [0.75,0.90]\boxplus 0.7\boxplus [0.75,0.90]}\\
 & =\frac{[0.39375,0.567]}{[3.2,3.5]}\\
 & =[0.1125,0.177188]
\end{align*}
\noindent
whence
\begin{align*}
x_{f_1,\pi_4}& =1\boxdot [0.1125,0.177188]\\
&=[0.1125,0.177188]\\
y_{f_2,\pi_4}& =[0.75,0.90]\boxdot [0.1125,0.177188]\\
&=[0.084375,0.159469]\\
z_{f_3,\pi_4}& =0.7\boxdot [0.1125,0.177188]\\
&=[0.07875,0.124031]\\
w_{I_t,\pi_4}&=[0.75,0.90]\boxdot [0.1125,0.177188]\\
&= [0.084375,0.159469]
\end{align*}
\end{itemize}

\begin{itemize}
\item The product $\pi_7=[0.007,0.04375]$ is distributed to $f_1$, $f_2$, $f_3$ and $I_t$ according to PCR6 as follows
\begin{align*}
 \frac{x_{f_1,\pi_7}}{1} & = \frac{y_{f_2,\pi_7}}{[0.10,0.25]} =  \frac{z_{f_3,\pi_7}}{0.7} = \frac{w_{I_t,\pi_7}}{[0.10,0.25]}\\
 & = \frac{\pi_7}{1\boxplus [0.10,0.25]\boxplus 0.7\boxplus [0.10,0.25]}\\
 & =\frac{[0.007,0.04375]}{[1.9,2.2]}\\
 & = [0.003182,0.023026]
\end{align*}
\noindent
whence
\begin{align*}
x_{f_1,\pi_7}& =1\boxdot [0.003182,0.023026]\\
&=[0.003182,0.023026]\\
y_{f_{2,\pi_7}}& =[0.10,0.25]\boxdot [0.003182,0.023026]\\
&=[0.000318,0.005757]\\
z_{f_3,\pi_7}& =0.7\boxdot [0.003182,0.023026]\\
&=[0.002227,0.016118]\\
w_{I_t,\pi_7}&=[0.10,0.25]\boxdot [0.003182,0.023026]\\
&=[0.000318,0.005757]
\end{align*}
\end{itemize}

\begin{itemize}
\item The product $\pi_8=[0.0525,0.1575]$ is distributed to $f_1$,  $f_3$ and $I_t$ according to PCR6 as follows
\begin{align*}
 \frac{x_{f_1,\pi_8}}{1} & =  \frac{z_{f_3,\pi_8}}{0.7} = \frac{w_{I_t,\pi_8}}{[0.10,0.25]\boxplus [0.75,0.90]}\\
  & = \frac{\pi_8}{1\boxplus 0.7\boxplus [0.10,0.25]\boxplus [0.75,0.90]}\\
 & =\frac{[0.0525,0.1575]}{[2.55,2.85]}\\
 & =[0.018421,0.061765]
\end{align*}
\noindent
whence
\begin{align*}
x_{f_1,\pi_8}& =1\boxdot [0.018421,0.061765]\\
&=[0.018421,0.061765]\\
z_{f_3,\pi_8}& =0.7\boxdot [0.018421,0.061765]\\
&=[0.012895,0.043235]\\
w_{I_t,\pi_8}&=[0.85,1.15]\boxdot [0.018421,0.061765]\\
&= [0.015658,0.071029]
\end{align*}
\end{itemize}

Summing the results, we get for $m_0\oplus m_1\oplus m_2\oplus m_3$ with PCR6 the following imprecise $m_{PCR6}$ bba:
\begin{align*}
m_{PCR6}(\theta_8)& =\pi_1 \boxplus  \pi_2 \boxplus  \pi_5 \\
&= [0.19425,0.32925]\\
m_{PCR6}(f_1)& =x_{f_1,\pi_3}\boxplus x_{f_1,\pi_4}\boxplus x_{f_1,\pi_7}\boxplus x_{f_1,\pi_8}\\
& =[0.152524,0.323743]\\
m_{PCR6}(f_2)& =y_{f_2,\pi_3}\boxplus y_{f_2,\pi_4}\boxplus y_{f_2,\pi_7}\\
& =[ 0.100351,0.236255]\\
m_{PCR6}(f_3)& =z_{f_3,\pi_3}\boxplus z_{f_3,\pi_4}\boxplus z_{f_3,\pi_7}\boxplus z_{f_3,\pi_8}\\
& =[0.106767,0.226620]\\
m_{PCR6}(I_t)& =w_{I_t,\pi_4}\boxplus w_{I_t,\pi_7}\boxplus w_{I_t,\pi_8}\\
& =[0.100351,0.236255]
\end{align*}

\noindent
where $\boxplus$, $\boxdot$ and $\boxslash$ operators (i.e. the addition, multiplication and division of imprecise values), and other operators on sets, were defined in \cite{DSmTBook1-3}, Vol. 1, p127--130 by
$$S_1\boxplus S_2=\{x|x=s_1+ s_2, s_1\in S_1, s_2\in S_2\}$$
$$S_1\boxdot S_2=\{x|x=s_1\cdot s_2, s_1\in S_1, s_2\in S_2\}$$
$$S_1\boxslash S_2=\{x|x=s_1/ s_2, s_1\in S_1, s_2\in S_2\}$$
\noindent
with
\begin{align*}
\inf(S_1\boxplus S_2)&=\inf(S_1)+\inf(S_2)\\
\sup(S_1\boxplus S_2)&=\sup(S_1)+\sup(S_2)
\end{align*}
\noindent
and
\begin{align*}
\inf(S_1\boxdot S_2)&=\inf(S_1)\cdot \inf(S_2)\\
\sup(S_1\boxdot S_2)&=\sup(S_1)\cdot \sup(S_2)
\end{align*}
\noindent
and
\begin{align*}
\inf(S_1\boxslash S_2)&=\inf(S_1)/\sup(S_2)\\
\sup(S_1\boxslash S_2)&=\sup(S_1)/\inf \sup(S_2)
\end{align*}
We have summarized the results in Table \ref{ResultTable9}. The left column of this table corresponds to the imprecise values of $m_{PCR6}$ based on exact calculus with operators on sets (i.e. the exact calculus with imprecision). The right column of this table ($m_{PCR6}^{approx}$) corresponds to the result obtained with non exact calculus based on  results given in Examples 1 and 2 in right columns of Tables \ref{ResultTable1} and \ref{ResultTable2}. This is what we call approximate results since they are not based on exact calculus with operators on sets. One shows an important differences between results in left and right columns which can make an impact on final decision process when working with imprecise bba's and we suggest to always use exact calculus (more complicated) instead of approximate calculus (more easier) in order to get the real imprecision on bba's values. Same approach can be done for combining imprecise bba's with PCR5 (not reported in this paper).\\

\begin{table}[!h]
 \small
\centering
 \begin{tabular}{|l|c|c|}
    \hline
    focal element     & $m_{PCR6}(.)$  & $m_{PCR6}^{approx}(.)$ \\
    \hline  
    $\theta_8$ & [0.194250,0.329250] &  [0.24375,0.27300] \\
    $f_1=\theta_4\cup\theta_8$ & [0.152524,0.323743] & [0.23935,0.29641] \\    
    $f_2=\theta_6\cup\theta_8$ & [0.100351,0.236255] &  [0.14587, 0.16950]\\    
    $f_3=\overline{\theta_4\cup\theta_8}$ & [0.106767,0.226620] &  [0.14865,0.16811]\\
    $I_t$ & [0.100351,0.236255] & [0.14587, 0.16950] \\      
       \hline
  \end{tabular}
  \caption{Results of $m_0\oplus m_1\oplus m_2 \oplus m_3$ for Table \ref{Table9}.}
\label{ResultTable9}
\end{table}

Based on results on left column of Table \ref{ResultTable9}, one can easily compute the imprecise Bel and Pl values also of decision-support hypotheses which are
\begin{align*}
Bel(\theta_6\cup\theta_7\cup\theta_8)&=[0.294601,0.565505]\\
Pl(\theta_6\cup\theta_7\cup\theta_8)&=[0.654243,1.352123]\equiv [0.654243,1]\\
Bel(\theta_7\cup\theta_8)&=[0.194250,0.329250]\\
Pl(\theta_7\cup\theta_8)&=[0.654243,1.352123]\equiv [0.654243,1]\\
Bel(\theta_8)& =[0.194250,0.329250]\\
Pl(\theta_8)&=[0.547476,1.125502]\equiv [0.547476,1]
\end{align*}
\noindent

Therefore, one gets the following imprecision ranges for probabilities
\begin{align*}
P(\theta_6\cup\theta_7\cup\theta_8)&\in [0.294601,1]\\
P(\overline{\theta_6\cup\theta_7\cup\theta_8})&\in [0,0.705399]\\
P(\theta_7\cup\theta_8)&\in [0.194250,1]\\
P(\overline{\theta_7\cup\theta_8})&\in [0,0.805750]\\
P(\theta_8)&\in [0.194250,1]\\
P(\bar{\theta}_8)&\in [0,0.805750]
\end{align*}

Based on max of Bel or max of Pl criteria, one sees that the decision will be to evacuate the building $B$ whatever the decision-support hypothesis we prefer $\theta_6\cup\theta_7\cup\theta_8$, $\theta_7\cup\theta_8$ or $\theta_8$.\\

Let's compute now the imprecise DSmP values for $\epsilon=0.001$. The focal element $f_1=\theta_4\cup\theta_8$ is redistributed back to $\theta_4$ and $\theta_8$ directly proportionally to their corresponding masses and cardinalities
\begin{align*}
\frac{x_{\theta_4}}{0\boxplus 0.001} & = \frac{y'_{\theta_8}}{[0.194250,0.329250]\boxplus 0.001}\\
& = \frac{m_{PCR6}(\theta_4\cup\theta_8)}{0.002\boxplus [0.194250,0.329250]}\\
& = \frac{[0.152524,0.323743]}{[0.196250,0.331250]}\\
& = [\frac{0.152524}{0.331250},\frac{0.323743}{0.196250}]\\
& = [0.46045,1.64965]
\end{align*}

\noindent
whence
\begin{align*}
x_{\theta_4} & = 0.001\boxdot [0.46045,1.64965]\\
& = [0.000460,0.001650]\\
y'_{\theta_8} &= [0.195250,0.330250]\boxdot  [0.46045,1.64965]\\
&= [0.089903,0.544797]
\end{align*}

The focal element $f_2=\theta_6\cup\theta_8$ is redistributed back to $\theta_6$ and $\theta_8$ directly proportionally to their corresponding masses and cardinalities
\begin{align*}
\frac{z_{\theta_6}}{0\boxplus 0.001} & = \frac{y''_{\theta_8}}{[0.194250,0.329250]\boxplus 0.001}\\
& = \frac{m_{PCR6}(\theta_6\cup\theta_8)}{0.002\boxplus [0.194250,0.329250]}\\
& = \frac{[0.100351,0.236255]}{[0.196250,0.331250]}\\
& = [0.302943,1.20385]
\end{align*}

\noindent
whence
\begin{align*}
z_{\theta_6} & = 0.001\boxdot [0.302943,1.20385]\\
& = [0.000303,0.001204]\\
y''_{\theta_8} &= [0.195250,0.330250]\boxdot [0.302943,1.20385]\\
&= [0.0591496,0.3975714]
\end{align*}

\noindent
The focal element $f_3=\overline{\theta_4\cup\theta_8}$ which is also equal to $\theta_1\cup\theta_2\cup \theta_3\cup\theta_5\cup\theta_6\cup\theta_7$ is redistributed back to $\theta_1$, $\theta_2$, $\theta_3$, $\theta_5$, $\theta_6$ and $\theta_7$ directly proportionally to their corresponding masses and cardinalities
\begin{align*}
\frac{w_{\theta_1}}{0\boxplus 0.001} & = \frac{w_{\theta_2}}{0\boxplus 0.001} =\frac{w_{\theta_3}}{0\boxplus 0.001}=\frac{w_{\theta_5}}{0\boxplus 0.001} \\
& =  \frac{w_{\theta_6}}{0\boxplus 0.001} =\frac{w_{\theta_7}}{0\boxplus 0.001}\\
& = \frac{m_{PCR6}(\overline{\theta_4\cup\theta_8})}{0.006}\\
& = \frac{[0.106767,0.226620]}{0.006}\\
& = [17.7945,37.770]
\end{align*}

\noindent
Since all are equal, we get
\begin{align*}
w_{\theta_1} & = w_{\theta_2} = w_{\theta_3} =w_{\theta_5}=w_{\theta_6} =w_{\theta_7}\\
&= 0.001 \boxdot [17.7945,37.770]\\
&= [0.0177945,0.03777]
\end{align*}

\noindent
The total ignorance $I_t=\theta_1\cup\theta_2\cup \theta_3\cup \theta_4\cup \theta_5\cup\theta_6\cup\theta_7\cup  \theta_8$ is redistributed back to all eight elements of the frame $\Theta$ directly proportionally to their corresponding masses and cardinalities
\begin{align*}
\frac{v_{\theta_1}}{0\boxplus 0.001} & = \frac{v_{\theta_2}}{0\boxplus 0.001} =\frac{v_{\theta_3}}{0\boxplus 0.001}=\frac{v_{\theta_4}}{0\boxplus 0.001} \\
& =  \frac{v_{\theta_5}}{0\boxplus 0.001}  = \frac{v_{\theta_6}}{0\boxplus 0.001} =\frac{v_{\theta_7}}{0\boxplus 0.001}\\
& =  \frac{v_{\theta_8}}{[0.194250,0.329250]\boxplus 0.001} \\
& = \frac{m_{PCR6}(I_t)}{[0.194250,0.329250]\boxplus 0.008}\\
& = \frac{[0.100351,0.236255]}{[0.202550,0.337250}\\
& = [0.297557,1.16813]
\end{align*}

\noindent
whence
\begin{align*}
v_{\theta_1} & = v_{\theta_2} = v_{\theta_3} =v_{\theta_4}=v_{\theta_5}=v_{\theta_6} =v_{\theta_7}\\
&= 0.001 \boxdot [0.297557,1.16813]\\
&= [0.000298,0.001168]\\
v_{\theta_8} & = [0.195250,0.330250]\boxdot [0.297557,1.16813]\\
& =[0.058098,0.385775]
\end{align*}

The imprecise DSmP probabilities are computed by
\begin{align*}
DSmP_{\epsilon,PCR6}(\theta_1)&=w_{\theta_1} \boxplus v_{\theta_1} \\
DSmP_{\epsilon,PCR6}(\theta_2)&=w_{\theta_2} \boxplus v_{\theta_2}\\
DSmP_{\epsilon,PCR6}(\theta_3)&=w_{\theta_3} \boxplus v_{\theta_3} \\
DSmP_{\epsilon,PCR6}(\theta_4)&=x_{\theta_4}\boxplus v_{\theta_4} \\
DSmP_{\epsilon,PCR6}(\theta_5)&=w_{\theta_5} \boxplus v_{\theta_5} \\
DSmP_{\epsilon,PCR6}(\theta_6)&= z_{\theta_6}\boxplus w_{\theta_6} \boxplus v_{\theta_6} \\
DSmP_{\epsilon,PCR6}(\theta_7)&=w_{\theta_7} \boxplus v_{\theta_7} \\
DSmP_{\epsilon,PCR6}(\theta_8)&=y'_{\theta_8}\boxplus y''_{\theta_8} \boxplus v_{\theta_8}\
\end{align*}
    
\noindent
which are summarized\footnote{Actually for $\theta_8$, one gets with exact calculus of imprecision $DSmP_{\epsilon,PCR6}(\theta_8)=[0.2072,1.3281]$, but since a probability cannot be greater than 1, the upper bound of imprecision interval has been set to 1.}
 in Table \ref{ResultTable9DSmP} below.
 \begin{table}[!h]
 \small
\centering
 \begin{tabular}{|l|c|}
    \hline
 Singletons       & $DSmP_{\epsilon,PCR6}(.)$ \\
    \hline 
$\theta_1$ &      [0.0181,0.0389]  \\
$\theta_2$ &     [0.0181,0.0389]    \\
$\theta_3$ &      [0.0181,0.0389]   \\
$\theta_4$ &      [0.0008,0.0028] \\
$\theta_5$ &    [0.0181,0.0389]   \\
$\theta_6$ &    [0.0184    0.0402]   \\
$\theta_7$ &       [0.0181,0.0389]  \\
$\theta_8$ &       [0.2072,1]  \\
       \hline
  \end{tabular}
 \caption{Imprecise $DSmP_{\epsilon}$ of $m_0\oplus m_1\oplus m_2\oplus m_3$ for Table \ref{Table9}.}
\label{ResultTable9DSmP}
\end{table}       

\begin{itemize}
\item[-] {\bf{Answer to Q1:}} As we have shown, it is possible to fuse imprecise bba's with PCR6, and PCR5 too (see \cite{DSmTBook1-3}, Vol. 2) to get an imprecise result for decision-making support under uncertainty and imprecision. It is also possible to compute the exact imprecise values of DSmP if necessary. According to our analysis and our results, and using either the max of Bel, the max of Pl of the max of DSmP criterion, the decision will be to evacuate the building $B$. Of course, a similar analysis can be done to answer to the question Q2 when working with imprecise bba's, and for for computing imprecise BetP values as well.
\end{itemize}
  
\section{Qualitative approach}

\noindent
In this section we just show how the fusion and decision-making can be done using qualitative information expressed with labels. In our previous examples the quantitative baa's have been defined ad-hoc in satisfying some reasonable modeling and using minimal assumption compatible with what is given in the statement of the VBIED problem. The numerical values can be slightly changed (as we have shown in Examples 1 and 2, or in Examples 3 or 4) or they can even be taken as imprecise as in Example 9, but they still need to be kept coherent with sources reports in order to obtain what we consider as pertinent and motivated and fully justified answers to questions Q1 and Q2.

 In this section we show how to solve the problem using qualitative information using labels. We investigate the possibility to work either with a minimal set of labels $\{L_{1}=Low,L_{2}=High\}$ (i.e. with $m=2$ labels) or a more refined set consisting in 3 labels $\{L_{1}=Low,L_{2}=Medium,L_{3}=High\}$ (i.e. with $m=3$ labels). Each set is extended with minimal $L_0$ and maximal  $L_{m+1}$ labels as follows (see \cite{DSmTBook1-3}, Vol.3, Chap. 2 for examples and details)
$$\mathcal{L}_2=\{L_0\equiv 0, L_1=Low, L_2=High, L_3\equiv 1\}$$
\noindent
and
\begin{multline*}
\mathcal{L}_3=\{L_0\equiv 0, L_{1}=Low,L_{2}=Medium, \\
L_{3}=High, L_4\equiv 1\}
\end{multline*}

To simplify the presentation, we only present the results when combining directly the sources altogether and considering that they have all the same maximal reliability and importance in the fusion process. In other words, we just consider the qualitative counterpart of Example 1 only.

\subsection{Fusion of sources using $\mathcal{L}_2$}


\noindent
{\bf{Example 10}}:  When using  $\mathcal{L}_2$, the qualitative inputs\footnote{When dealing with qualitative information, we prefix the notations with 'q' letter, for example quantitative bba $m(.)$ becomes qualitative bba $qm(.)$, etc.} of the VBIED problem are chosen according to Table \ref{TableExample10}.

 \begin{table}[!h]
 \small
\centering
 \begin{tabular}{|l|c|c|c|c|}
    \hline
    focal element     & $qm_{0}(.)$  & $qm_{1}(.)$ & $qm_{2}(.)$ & $qm_{3}(.)$ \\
    \hline  
    $\theta_4\cup\theta_8$  & $L_3$ & $L_0$ & $L_1$  & $L_0$ \\
    $\theta_6\cup\theta_8$ & $L_0$ & $L_2$ &  $L_0$ & $L_1$\\
    $\overline{\theta_4\cup\theta_8}$  & $L_0$ & $L_0$ & $L_2$ & $L_0$\\
    $I_t$ & $L_0$ & $L_1$ & $L_0$ & $L_2$\\
   \hline
  \end{tabular}
  \caption{Qualitative inputs using $\mathcal{L}_2$.}
\label{TableExample10}
\end{table}

Using DSm field and linear algebra of refined labels based on equidistant labels assumption, one gets the following mapping  between labels and numbers $L_0\equiv 0$,  $L_1\equiv 1/3$, $L_2\equiv 2/3$ and $L_3\equiv 1$ and therefore, the Table \ref{TableExample10} is equivalent to the quantitative inputs table \ref{TableExample10quant} (which are close to the numerical values taken in Example 1).
 \begin{table}[!h]
 \small
\centering
 \begin{tabular}{|l|c|c|c|c|}
    \hline
    focal element     & $m_{0}(.)$  & $m_{1}(.)$ & $m_{2}(.)$ & $m_{3}(.)$ \\
    \hline  
    $\theta_4\cup\theta_8$  & $1$ & 0 & $1/3$  & 0 \\
    $\theta_6\cup\theta_8$ & 0 & $2/3$ & 0 & $1/3$\\
    $\overline{\theta_4\cup\theta_8}$  & 0 & 0 & $2/3$ & 0\\
    $I_t$ & 0 & $1/3$ & 0 & $2/3$\\
   \hline
  \end{tabular}
  \caption{Corresponding quantitative inputs.}
\label{TableExample10quant}
\end{table}

\noindent
Applying PCR5 and PCR6 fusion rules, one gets the results given in Table \ref{ResultTable10} for quantitative and approximate qualitative bba's.

 \begin{table}[!h]
 \small
\centering
 \begin{tabular}{|l|c|c|}
    \hline
    focal element     & $m_{PCR5}\approx qm_{PCR5}$  & $qm_{PCR6}\approx qm_{PCR6}$ \\
    \hline  
      $\theta_8$ & $0.25926\approx L_1$ & $0.25926\approx L_1$ \\
       $\theta_4\cup\theta_8$ & $0.36145\approx L_1$ & $0.3157 \approx L_1$ \\    
       $\theta_6\cup\theta_8$ & $0.093855\approx L_0$ &  $0.13198\approx L_0$\\    
        $\overline{\theta_4\cup\theta_8}$ & $0.19158\approx L_1$ &  $0.16108\approx L_0$\\
           $I_t$ & $0.093855\approx L_0$ & $0.13198\approx L_0$ \\      
       \hline
  \end{tabular}
  \caption{Results of $m_0\oplus m_1\oplus m_2 \oplus m_3$ for Table \ref{TableExample10}.}
\label{ResultTable10}
\end{table}

One sees that the crude approximation of numerical values to their closest corresponding labels in $\mathcal{L}_2$ can yield to unnormalized qualitative bba. For example, $qm_{PCR6}(.)$ is not normalized since the sum of labels of focal elements in the right column of Table \ref{ResultTable10} is $  L_1+ L_1+ L_0+ L_0+ L_0= L_2\neq L_3$. To preserve the normalization of qbba result it is better to work with refined labels as suggested in \cite{DSmTBook1-3}, Vol.3, Chap. 2. Using refined labels, one will get now a better approximation as shown in the Table \ref{ResultTable10refined}.

 \begin{table}[!h]
 \small
\centering
 \begin{tabular}{|l|c|c|}
    \hline
    focal element     & $m_{PCR5}\approx qm_{PCR5}$  & $m_{PCR6}\approx qm_{PCR6}$ \\
    \hline  
      $\theta_8$ & $0.25926\approx L_{0.79}$ & $0.25926\approx L_{0.79}$ \\
       $\theta_4\cup\theta_8$ & $0.36145\approx L_{1.08}$ & $0.31570 \approx L_{0.95}$ \\    
       $\theta_6\cup\theta_8$ & $0.09385\approx L_{0.28}$ &  $0.13198\approx L_{0.39}$\\    
        $\overline{\theta_4\cup\theta_8}$ & $0.19158\approx L_{0.57}$ &  $0.16108\approx L_{0.48}$\\
           $I_t$ & $0.09385\approx L_{0.28}$ & $0.13198\approx L_{0.39}$ \\      
       \hline
  \end{tabular}
  \caption{Results of $m_0\oplus m_1\oplus m_2 \oplus m_3$ with refined labels.}
\label{ResultTable10refined}
\end{table}
It can be easily verified that the qbba's based on refined label approximations are now (qualitatively) normalized (because the sum of refined labels of each column is equal to $L_3$).\\

 The results of qDSmP based on refined and crude approximations are given in Table \ref{ResultTableExample10DSmP}.   
    \begin{table}[!h]
 \small
\centering
 \begin{tabular}{|l|c|c|}
    \hline
 Singletons       & $qDSmP_{\epsilon,PCR5}(.)$  & $qDSmP_{\epsilon,PCR6}(.)$ \\
    \hline 
$\theta_1$ &     $0.0323 \approx L_{0.10} \approx L_0$  &  $0.0273 \approx L_{0.08} \approx L_0$\\
$\theta_2$ &     $0.0323 \approx L_{0.10}\approx L_0$  &  $0.0273\approx L_{0.08}\approx L_0$\\
$\theta_3$ &      $0.0323 \approx L_{0.10}\approx L_0$  &  $0.0273\approx L_{0.08}\approx L_0$\\
$\theta_4$ &      $0.0017 \approx L_{0.00}\approx L_0$  &    $0.0017\approx L_{0.01}\approx L_0$\\
$\theta_5$ &      $0.0323 \approx L_{0.10}\approx L_0$  &  $0.0274\approx L_{0.08}\approx L_0$\\
$\theta_6$ &      $0.0326 \approx L_{0.10}\approx L_0$  &  $0.0279\approx L_{0.08}\approx L_0$\\
$\theta_7$ &      $0.0323\approx L_{0.10}\approx L_0$  &   $ 0.0273\approx L_{0.08}\approx L_0$\\
$\theta_8$ &      $0.8042 \approx L_{2.40}\approx L_2$  &   $0.8338\approx L_{2.51}\approx L_3$\\
       \hline
  \end{tabular}
  \caption{Results of $qDSmP_{\epsilon}$ for Table \ref{TableExample10}.}
\label{ResultTableExample10DSmP}
\end{table}    

\noindent
{\bf{Answer to Q1 using crude approximation:}} Based on these qualitative results, one sees that using crude approximation (i.e. using only labels in $\mathcal{L}_2$)  one gets\footnote{The derivations of $qBel(\bar{X})$ and $qPl(\bar{X})$ were obtained using qualitative extension of Dempster's formulas \cite{Shafer_1976}, i.e. with $qBel(\bar{X})=L_m-qPl(X)$ and $qPl(\bar{X})=L_m-qBel(X)$. These results are valid only if the qbba is normalized, but are used here even when using non normalized qbba as crude approximation.}

\begin{itemize}
\item with qPCR5
\begin{align*}
qP(\theta_6\cup\theta_7\cup\theta_8)&\in [ L_1, L_3]\\
qP(\overline{\theta_6\cup\theta_7\cup\theta_8})&\in [ L_0, L_2]\\
qP(\theta_7\cup\theta_8)&\in [ L_1, L_3]\\
qP(\overline{\theta_7\cup\theta_8})&\in [ L_0, L_2]\\
qP(\theta_8)&\in [ L_1, L_2]\\
qP(\bar{\theta}_8)&\in [ L_1, L_2]
\end{align*}
\item with qPCR6
\begin{align*}
qP(\theta_6\cup\theta_7\cup\theta_8)&\in [ L_1, L_2]\\
qP(\overline{\theta_6\cup\theta_7\cup\theta_8})&\in [ L_1, L_2]\\
qP(\theta_7\cup\theta_8)&\in [ L_1, L_2]\\
qP(\overline{\theta_7\cup\theta_8})&\in [ L_1, L_2]\\
qP(\theta_8)&\in [ L_1, L_2]\\
qP(\bar{\theta}_8)&\in [ L_1, L_2]
\end{align*}
\end{itemize}
These results show that is is almost impossible to answer clearly and fairly to the question Q1 using the max of Bel or the max of Pl criteria based on such very inaccurate qualitative bba's using such crude approximation. However it is possible and easy to answer to Q1 using qualitative DSmP value. However and according to Table \ref{ResultTableExample10DSmP}, the final decision must be to {\bf{evacuate the building $B$}} when considering the level of DSmP values of $\theta_6\cup\theta_7\cup\theta_8$, $\theta_7\cup\theta_8$, or $\theta_8$. \\

\noindent
{\bf{Answer to Q1 using refined approximation:}} Using the refined approximation using refined labels which is more accurate, one gets
\begin{itemize}
\item with qPCR5
\begin{align*}
qP(\theta_6\cup\theta_7\cup\theta_8)&\in [ L_{1.07}, L_3]\\
qP(\overline{\theta_6\cup\theta_7\cup\theta_8})&\in [ L_0, L_{1.93}]\\
qP(\theta_7\cup\theta_8)&\in [ L_{0.79}, L_3]\\
qP(\overline{\theta_7\cup\theta_8})&\in [ L_{0},  L_{2.21}]\\
qP(\theta_8)&\in [ L_{0.79}, L_{2.43}]\\
qP(\bar{\theta}_8)&\in [ L_{0.57}, L_{2.21}]
\end{align*}
\item with qPCR6
\begin{align*}
qP(\theta_6\cup\theta_7\cup\theta_8)&\in [ L_{1.18}, L_3]\\
qP(\overline{\theta_6\cup\theta_7\cup\theta_8})&\in [ L_0, L_{1.82}]\\
qP(\theta_7\cup\theta_8)&\in [ L_{0.79}, L_3]\\
qP(\overline{\theta_7\cup\theta_8})&\in [ L_{0},  L_{2.21}]\\
qP(\theta_8)&\in [L_{0.79}, L_{2.52}]\\
qP(\bar{\theta}_8)&\in [ L_{0.48}, L_{2.21}]
\end{align*}
\end{itemize}
One sees that accuracy of the result obtained using refined labels allows us to take the decision more easily. Indeed, using the refined approximation, it is possible here to take  the decision based on the max of Bel, or on the max of Pl and whatever the decision-support hypothesis used ($\theta_6\cup\theta_7\cup\theta_8$, or $\theta_7\cup\theta_8$, or $\theta_8$), the answer to question Q1 is: {\bf{Evacuation of the building $B$}}. The same decision can also be taken from the analysis of qDSmP values as well when considering refined labels in Table \ref{ResultTableExample10DSmP}.

\subsection{Fusion of sources using $\mathcal{L}_3$}
    
Here we propose to go further in our analysis and to use a bit more refined set of labels defined by $\mathcal{L}_3$.
We need to adapt the qualitative inputs of the VBIED problem in order to work with $\mathcal{L}_3$.\\


\noindent
{\bf{Example 11}}:  We propose to solve the VBIED problem for the following qualitative inputs which reflects what is reported by the sources when using labels belonging to $\mathcal{L}_3$.

 \begin{table}[!h]
 \small
\centering
 \begin{tabular}{|l|c|c|c|c|}
    \hline
    focal element     & $qm_{0}(.)$  & $qm_{1}(.)$ & $qm_{2}(.)$ & $qm_{3}(.)$ \\
    \hline  
    $\theta_4\cup\theta_8$        & $L_4$ & $L_0$          & $L_1$  & $L_0$ \\
    $\theta_6\cup\theta_8$       & $L_0$          & $L_3$ & $L_0$          & $L_1$\\
    $\overline{\theta_4\cup\theta_8}$        & $L_0$          & $L_0$          & $L_3$  & $L_0$\\
    $I_t$ & $L_0$         & $L_1$ & $L_0$            & $L_3$\\
   \hline
  \end{tabular}
  \caption{Qualitative inputs based on $\mathcal{L}_3$.}
\label{TableExample11}
\end{table}

Based on the equidistant labels assumption, one gets the following mapping  between labels and numbers $L_0\equiv 0$,  $L_1\equiv 1/4$, $L_2\equiv 2/4$, $L_3\equiv 3/4$ and $L_4=1$ and therefore, the Table \ref{TableExample11} is equivalent to the quantitative inputs table \ref{TableExample11quant} (which are more close to the numerical values taken in Example 1 than the inputs chosen in Table \ref{TableExample10quant} for Example 10).

 \begin{table}[!h]
 \small
\centering
 \begin{tabular}{|l|c|c|c|c|}
    \hline
    focal element     & $m_{0}(.)$  & $m_{1}(.)$ & $m_{2}(.)$ & $m_{3}(.)$ \\
    \hline  
    $\theta_4\cup\theta_8$        & 1 & 0          & 0.25  & 0 \\
    $\theta_6\cup\theta_8$       & 0          & 0.75 & 0           & 0.25\\
    $\overline{\theta_4\cup\theta_8}$        & 0          & 0          & 0.75  & 0\\
    $I_t$ & 0         & 0.25 & 0            & 0.75\\
   \hline
  \end{tabular}
  \caption{Corresponding quantitative inputs.}
\label{TableExample11quant}
\end{table}

\noindent
Applying PCR5 and PCR6 fusion rules, one gets the results given in Table \ref{ResultTable11} for quantitative and approximate qualitative bba's using refined and crude approximations of labels.
 \begin{table}[!h]
\tiny
\centering
 \begin{tabular}{|l|c|c|}
    \hline
    foc. elem.    & $m_{PCR5}\approx qm_{PCR5}$  & $m_{PCR6}\approx qm_{PCR6}$ \\
    \hline  
      $\theta_8$ & $0.20312 \approx L_{0.81} \approx L_1$ & $0.20312\approx L_{0.81}\approx L_1$ \\
       $\theta_4\cup\theta_8$ & $0.34269\approx L_{1.37} \approx L_1$ & $0.29979 \approx L_{1.21} \approx L_1$ \\    
       $\theta_6\cup\theta_8$ & $0.11617\approx L_{0.47}\approx L_0$ &  $0.15370\approx L_{0.61}\approx L_1$\\  
        $\overline{\theta_4\cup\theta_8}$ & $0.22185\approx L_{0.88} \approx L_1$ &  $0.18969\approx L_{0.76}\approx L_1$\\  
           $I_t$ & $0.11617\approx L_{0.47}\approx L_0$ & $0.15370\approx L_{0.61}\approx L_1$ \\      
       \hline
  \end{tabular}
 \caption{Results of $m_0\oplus m_1\oplus m_2 \oplus m_3$ for Table \ref{TableExample11}.}
\label{ResultTable11}
\end{table}

It can be easily verified that the qbba's based on refined label approximations are (qualitatively) normalized because the sum of refined labels of each column is equal to $L_4$. Using crude approximation when working only with labels in $\mathcal{L}_3$ we get non normalized qbba's. The results of qDSmP based on refined and crude approximations are given in Table \ref{ResultTableExample11DSmP}.   
   
    \begin{table}[!h]
 \tiny
\centering
 \begin{tabular}{|l|c|c|}
    \hline
 Singletons       & $qDSmP_{\epsilon,PCR5}(.)$  & $qDSmP_{\epsilon,PCR6}(.)$ \\
    \hline 
$\theta_1$ &     $0.0375 \approx L_{0.15}\approx L_0$  &  $0.0323\approx L_{0.13}\approx L_0$\\
$\theta_2$ &     $0.0375  \approx L_{0.15}\approx L_0$  &  $0.0323\approx L_{0.13}\approx L_0$\\
$\theta_3$ &      $0.0375  \approx L_{0.15}\approx L_0$  &  $0.0323\approx L_{0.13}\approx L_0$\\
$\theta_4$ &      $0.0022  \approx L_{0.01}\approx L_0$  &    $0.0022\approx L_{0.01}\approx L_0$\\
$\theta_5$ &      $0.0375  \approx L_{0.15}\approx L_0$  &  $0.0323\approx L_{0.13}\approx L_0$\\
$\theta_6$ &      $0.0381  \approx L_{0.15}\approx L_0$  &  $0.0331\approx L_{0.13}\approx L_0$\\
$\theta_7$ &      $0.0375  \approx L_{0.15}\approx L_0$  &   $ 0.0323\approx L_{0.13}\approx L_0$\\
$\theta_8$ &      $0.7722  \approx L_{3.09}\approx L_3$  &   $0.8032\approx L_{3.21}\approx L_3$\\
       \hline
  \end{tabular}
  \caption{Results obtained with $qDSmP_{\epsilon}$ for Table \ref{TableExample11}.}
\label{ResultTableExample11DSmP}
\end{table}    

Ones sees that the use of refined labels allows to obtain normalized qualitative probabilities. This is not possible to get normalized qualitative probabilities when using only crude approximations with labels in $\mathcal{L}_3$ for this example.\\

\noindent
{\bf{Answer to Q1:}} Using refined labels (which is more accurate), one gets finally
\begin{itemize}
\item with qPCR5
\begin{align*}
qP(\theta_6\cup\theta_7\cup\theta_8)&\in [ L_{1.28}, L_4]\\
qP(\overline{\theta_6\cup\theta_7\cup\theta_8})&\in [ L_{0},  L_{2.72}]\\
qP(\theta_7\cup\theta_8)&\in [ L_{0.81}, L_4]\\
qP(\overline{\theta_7\cup\theta_8})&\in [ L_{0}, L_{3.19}]\\
qP(\theta_8)&\in [ L_{0.81}, L_{3.12}]\\
qP(\bar{\theta}_8)&\in [ L_{0.88}, L_{3.19}]
\end{align*}
\item with qPCR6
\begin{align*}
qP(\theta_6\cup\theta_7\cup\theta_8)&\in [ L_{1.42}, L_4]\\
qP(\overline{\theta_6\cup\theta_7\cup\theta_8})&\in [ L_{0},  L_{2.58}]\\
qP(\theta_7\cup\theta_8)&\in [ L_{0.81}, L_4]\\
qP(\overline{\theta_7\cup\theta_8})&\in [ L_{0},  L_{3.19}]\\
qP(\theta_8)&\in [L_{0.81}, L_{3.24}]\\
qP(\bar{\theta}_8)&\in [ L_{0.76}, L_{3.19}]
\end{align*}
\end{itemize}

One sees that based with PCR5 or PCR6 whatever the decision-support hypothesis we consider ($\theta_6\cup\theta_7\cup\theta_8$, $\theta_7\cup\theta_8$, or $\theta_8$), one will decide to evacuate the building $B$ based on max of Bel, max of Pl or DSmP values, except for the case of PCR5 with $\theta_8$ based on the max of Bel, max of Pl. In this case, PCR5 result suggests to NOT evacuate $B$ contrariwise to PCR6 result. As far as $\theta_8$ is the preferred (optimistic) decision-support hypothesis, one sees here the main effect of difference between PCR5 and PCR6 for decision-making support. But as already stated, for such problem the most prudent strategy for decision-making is to consider the decisio-support hypothesis $\theta_6\cup\theta_7\cup\theta_8$ which captures all aspects of potential danger. Using such reasonable strategy, both rules PCR5 and PCR6 yields same decision: Evacuation of the building $B$.

\section{Conclusions}

In this paper we have presented a modeling for solving the Vehicle-Born Improvised Explosive Device (VBIED) problem with Dezert-Smarandache Theory (DSmT) framework. We have shown how it is possible to compute imprecise probabilities of all decision-support hypotheses and how to take into account the reliabilities and the importances of the sources of information in decision-making support.
The strong impact of prior information has also been analyzed, as well as the possibility to deal directly with imprecise sources of information and even with qualitative reports. We have answered with the full justification to the two main questions asked in the VBIED problem by John Lavery and Simon Maskell: 1) what is  the final decision to take, and 2) what is the best fusion subsystem to choose (APNR or the pool of experts)? The analysis done in this paper is based on a very limited number of reasonable assumptions and could be adapted for solving more complicated security problems involving imprecise, incomplete and conflicting sources of information.


\begin{thebibliography}{99}


\bibitem{Dambreville_2010}
F. Dambreville, \emph{Generic implementation of fusion rules based on Referee function}, Proc. of Workshop on the theory of belief functions, April 1-2, 2010, Brest, France.

\bibitem{Dezert_2010b}
J. Dezert, J.-M. Tacnet, M. Batton-Hubert, F. Smarandache, \emph{Multi-criteria decision making based on DSmT-AHP}, Int. Workshop on Belief Functions, Brest, France, April 2010.

\bibitem{MartinDSmTBook2}
A.~Martin, C. Osswald, \emph{A new generalization of the proportional conflict redistribution rule stable in terms of decision}, Chapter 2 in \cite{DSmTBook1-3}, Vol. 2, 2006.


 \bibitem{MartinDSmTBook3}
A.~Martin, \emph{Implementing general belief function framework with a practical codification for low complexity}, Chapter 7 in \cite{DSmTBook1-3}, Vol. 3, 2009.

\bibitem{Shafer_1976}
G. Shafer, \emph{A mathematical theory of evidence}, Princeton University Press, 1976.

 \bibitem{DSmTBook1-3}
F.~Smarandache, J.~Dezert, \emph{Advances and applications of DSmT for information fusion (Collected works)}, Vols. 1-3, American Research Press, 2004--2009.
{\tiny{http://www.gallup.unm.edu/{\verb+~+}smarandache/DSmT.htm}}

\bibitem{FSJDJMT2010}
F. Smarandache, J. Dezert, J.-M. Tacnet, \emph{Fusion of sources of evidence with different importances and reliabilities}, accepted in Fusion 2010 conf, Edinburgh, July 26-29, 2010.

 \bibitem{Smets1990}
Ph. Smets Ph., \emph{The Combination of Evidence in the Transferable Belief Model}, IEEE Trans. PAMI 12, pp. 447--458, 1990.

 \bibitem{Sudano2002}
J. Sudano, \emph{The system probability information content (PIC) relationship to contributing components, combining independent multi-source beliefs, hybrid and pedigree pignistic probabilities}, Proc. of Fusion 2002, Vol.2, pp. 1277-1283, Annapolis, MD, USA, July 2002.

\end{thebibliography}
\end{document}